\documentclass[sigconf]{aamas}

\usepackage{balance} 
\usepackage{dblfloatfix}
\usepackage{graphicx}
\usepackage{subcaption}
\usepackage{gensymb}
\usepackage{booktabs}
\usepackage{xcolor}
\usepackage{algorithm}
\usepackage{amsfonts}
\usepackage{amsmath}
\usepackage{algpseudocode}
\usepackage{courier}

\newcommand{\interpretation}{I}

\def\nearport{\textsf{nearport}}
\def\hotspot{\textsf{hotspot}}
\def\aisoff{\textsf{ais-off}}
\def\stay{\textsf{stay}}
\def\draught{\textsf{draught}}
\def\highspeed{\textsf{high-speed}}
\def\lowspeed{\textsf{low-speed}}
\def\changedirection{\textsf{change-direction}}
\def\parsimony{\sigma}

\newtheorem{example}{Example}[section]

\newcommand{\constantSet}{\mathcal{C}}

\newcommand{\predicateSet}{\mathcal{P}}
\newcommand{\variableSet}{\mathcal{V}}

\newcommand{\groundLiteralSet}{\mathcal{G}}
\newcommand{\interpretationSet}{\mathcal{I}}
\newcommand{\semiLattice}{\mathcal{L}}

\newcommand{\satisfactionAtTime}[1]{\models_{#1}}
\newcommand{\program}{\Pi}

\newcommand{\groundedLiteral}{g}

\newcommand{\trajectory}{\tau}
\newcommand{\trajectoryset}{\mathrm{T}}
\def\domainSet{\mathcal{D}}

\def\domSet{\domainSet}
\def\atPred{\textsf{at}}

\def\regionconst{\textsf{r}}

\def\after{\textsf{AFTER}}

\newtheorem{definition}{Definition}

\def\avar{\textsf{AVar}}

\makeatletter
\gdef\@copyrightpermission{
  \begin{minipage}{0.2\columnwidth}
   \href{https://creativecommons.org/licenses/by/4.0/}{\includegraphics[width=0.90\textwidth]{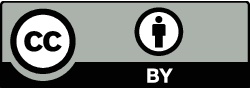}}
  \end{minipage}\hfill
  \begin{minipage}{0.8\columnwidth}
   \href{https://creativecommons.org/licenses/by/4.0/}{This work is licensed under a Creative Commons Attribution International 4.0 License.}
  \end{minipage}
  \vspace{5pt}
}
\makeatother

\setcopyright{ifaamas}
\acmConference[AAMAS '25]{Proc.\@ of the 24th International Conference
on Autonomous Agents and Multiagent Systems (AAMAS 2025)}{May 19 -- 23, 2025}
{Detroit, Michigan, USA}{Y.~Vorobeychik, S.~Das, A.~Nowé  (eds.)}
\copyrightyear{2025}
\acmYear{2025}
\acmDOI{}
\acmPrice{}
\acmISBN{}

\acmSubmissionID{<<OpenReview submission id>>}

\title[Maritime Abduction for Region Generation]{Sea-cret Agents: Maritime Abduction for Region Generation to Expose Dark Vessel Trajectories}

\author{Divyagna Bavikadi}
\affiliation{
  \institution{Arizona State Universty}
  \city{Tempe}
  \country{USA}}
\email{dbavikad @asu.edu}

\author{Nathaniel Lee}
\affiliation{
  \institution{Arizona State Universty}
  \city{Tempe}
  \country{USA}}
\email{nlee51@asu.edu}

\author{Paulo Shakarian}
\affiliation{
  \institution{Arizona State Universty}
  \city{Tempe}
  \country{USA}}
\email{pshakari@asu.edu}

\author{Chad Parvis}
\affiliation{
  \institution{EpochGeo}
  \city{Washington, DC}
  \country{USA}}
\email{cp@epochgeo.com}

\begin{abstract}
Bad actors in the maritime industry engage in illegal behaviors after disabling their vessel's automatic identification system (AIS) - which makes finding such vessels difficult for analysts.  Machine learning approaches only succeed in identifying the locations of these ``dark vessels'' in the immediate future.  This work leverages ideas from the literature on abductive inference applied to locating adversarial agents to solve the problem.  Specifically, we combine concepts from abduction, logic programming, and rule learning to create an efficient method that approaches full recall of dark vessels while requiring less search area than machine learning methods.  We provide a logic-based paradigm for reasoning about maritime vessels, an abductive inference query method, an automatically extracted rule-based behavior model methodology, and a thorough suite of experiments.
\end{abstract}

\keywords{Maritime forecasting; Abductive Reasoning; Agentic Behavior.}

\newcommand{\BibTeX}{\rm B\kern-.05em{\sc i\kern-.025em b}\kern-.08em\TeX}

\begin{document}


\pagestyle{fancy}
\fancyhead{}


\maketitle 

\section{Introduction}
Maritime vessels are equipped with an automatic identification system (AIS) to track their position on the globe~\cite{ais}.  However, malicious actors often disable this system - becoming ``dark'' when conducting illegal activities.  Understanding these ``dark vessels'' has implications for security~\cite{darkvesselsecurity}, maritime analysis~\cite{maritime-trajectory-mining, anomaly-detection}, planning~\cite{traffic}, and forecasting~\cite{forecasting}.  Recently, with the support of the U.S. Treasury and European Union in enforcing maritime services prohibitions for seaborne Russian oil~\cite{wsj1}, industry efforts have targeted real-world issues such as illegal fishing, human trafficking, border protection, and sanction violations ~\cite{spire2024, windward2024}, highlighting the increasing need for efficient dark vessel detection.  Recent machine learning (ML) approaches are limited to trajectory prediction with a time horizon of less than an hour~\cite{tf,tp,tf2,groupwaterarea} or rely on satellite data susceptible to weather conditions~\cite{capella, satellite, canada}.  Other approaches require expert intervention using the radio frequency Doppler shift~\cite{spire2024}. These approaches are not data-efficient and cannot explain why they determined a given result.  We note that from a practical perspective, the limited forward-prediction value of ML approaches is significant - but the ability to find the dark vessel locations degrades with increased search area and resources.  Meanwhile, recent work on generating faux trajectories for human movement suggests that abductive inference can address some of these difficulties~\cite{bavikadi2024geospatial} - although that work does not predict real trajectories and was not applied to the maritime domain.  In this paper, we combine ideas from abductive inference, logic programming, and rule learning to identify the locations of dark vessels based on partial trajectories.  We show that we are able to approach full recall of dark vessel trajectories requiring less than half of the area coverage required by our machine learning baselines. Further, we found that the recall performance of the abduction-based approach \textit{increases} with search area and resources - unlike the degradation experienced with ML.  We also demonstrate data efficiency, efficient inference calculations, and describe our ongoing efforts to deploy this technology in an operational platform.  After a review of background material (Section~\ref{sec:background}) we make the following contributions:

\begin{enumerate}
\vspace{-6pt}
    \item We provide a formalism for reasoning about maritime vessels including a logical language to express maritime vessel trajectories (Section~\ref{subsec:prelims}) and the framing of an abduction problem (Section~\ref{sec:abduc}) that include a top-$k$ approximation that we explore empirically in this paper.
    \item We provide a simple but effective rule-learning approach to agent behavior modeling (Section~\ref{sec:rule_learning}) that not only allows for data-driven (and data-efficient) abduction but also affords explainability of the results.
    \item We provide a suite of experimental results (Section~\ref{sec:exp}) that demonstrate how the abduction approach is area-efficient by saturating with $157\%$ higher recall than baselines for an area of $30 km^2$, provides long-term predictions where ML methods fail, and provides improved performance of $476\%$ in recall with additional resources.  
    \item We also show that the approach is efficient in both terms of runtime and data as it can be instantiated with very little data - even a single training trajectory (providing comparable performance of $0.62$ precision to the use of all historical data - where we show ML catastrophically fails), as well as provide various ablation studies.
    \item We describe our efforts to deploy this system on an operational platform to support real-world analysts in the discovery of dark vessels (Section~\ref{sec:deploy}).
\end{enumerate}

\section{Background}
\label{sec:background}
\noindent\textbf{Dark Vessel Analytics.}
Maritime vessels employ deceptive shipping practices to benefit from violating international law, conducting illicit operations, violating environmental protections, and avoiding sanctions. In the 18th century, vessels disguised their Jolly Roger flags to deceive prospective victims before attacking them. Currently, vessels manipulate their AIS to avoid being monitored while performing illicit activities. On a monthly average, $800,000$ dark activity events were detected in $2020-2022$~\cite{windward2024}. Lately, in the aftermath of the Russia-Ukraine war, sanctions on maritime trade have evolved~\cite{wsj1}, and monthly dark activity rose by $216\%$~\cite{wind24}. More recently, in $2024$, there has been a $340\%$ rise in dark activity~\cite{dark2024} much of which focused on the Black Sea (the area used in our experiments in Section~\ref{sec:exp}). Such activities when gone undetected, can have realistic detrimental impacts on ecosystems, safety, trade, and security. To monitor and control such behavior, efforts from government agencies~\cite{wsj1, canada}, and industry~\cite{windward2024,capella,lockheed, spire2024}, have invested in various efforts that began in earnest with the DARPA PANDA program over a decade ago~\cite{panda}.  These programs have led to a line of research that we describe in the next subsection.

\vspace{2pt}

\noindent\textbf{Related Work.}  Earlier work on maritime vessel trajectory prediction relied on Markov models~\cite{markov3,markov4}, and extensions have also been applied to make efficient predictions. However, we find Hidden Markov Model to run out of memory (exceeding 200GB) during hidden-state extraction due to the trajectory size, a common step in traditional Markov approaches~\cite{memory1, memory2}. In contrast, our method efficiently handles a similar step - extracting region types without running out of memory. Markov models are known to result in reduced performance as the trajectory size increases~\cite{DLmove}. Also, Markov models work well for simple finite parameters but are unable to capture complex patterns and this led to the later use of deep learning techniques for the problem - further enabled by the availability of large datasets of maritime trajectories. To address the complexities of spatio-temporal interactions, \cite{dlbaseline} provides a sequence-to-sequence RNN to predict future maritime trajectories.  Related work looks to predict a point ship location using an LSTM-transformer combination~\cite{dstnet,txlstm}.  These methods differ from our approach as they only provide accurate predictions up to an hour in the future, require large amounts of training data, and do not afford explainability (so the analyst user cannot easily justify the dark vessel predictions to operational personnel).  Maritime trajectory patterns are also studied widely for traffic management~\cite{pattern-mining-2} with an unsupervised hierarchical method and safety~\cite{pattern-mining-3} where they mine patterns to focus on shipping route characterization and anomaly detection.  These methods are valuable for understanding typical and atypical trajectory behavior, but they primarily focus on identifying patterns after the fact. In contrast, our method leverages trajectory behavior through abductive reasoning to infer an agent's future locations. This work varies from other maritime applications of Artificial Intelligence (AI) like vessel detection~\cite{vesseldet} where a model generates bounding boxes for the object vessel in an image or tracking it in a video~\cite{tracking}. This work also differs from a complementary line of work of patrolling strategies~\cite{tambe, patrol1, patrol2} that generates optimal patrol locations to cover a set of targets as we focus on generating locations to capture a target at a time horizon (as opposed to developing patrol plans for a non-adversarial agent). Trajectory forecasting is a separate line of work, it is focused on short-time horizon prediction of human or robotic movement as opposed to the long-time horizon, global-scale prediction of maritime vessels.  Some notable approaches use deep learning architectures based on convolutional networks~\cite{tf}, adversarial methods~\cite{tp}, autoencoders~\cite{tf2}, and Markov chains~\cite{markov1,markov2}.

Abductive inference has provided a natural paradigm for locating unobserved adversarial agents - requiring much less data and providing more transparency than ML methods.  Early work in this area offered simple models relating the adversary's point location to geospatial phenomenon~\cite{DBLP:journals/tist/ShakarianSS11}.  Later work took a data-driven approach to learn a model of the adversarial behavior that enables abductive inference~\cite{DBLP:conf/cikm/ShaabaniASS16}.  None of the aforementioned prior work on abduction involves trajectories nor does it involve making predictions of agent behavior over a long time horizon.  Complementary to abduction work is the generation of spatial regions~\cite{regiongen}, which aims to maintain meaningful spatial boundaries for transportation services by partitioning an area of interest via region clustering (we employ similar techniques during pre-processing).  More recent work on abductive inference has been applied to human movement~\cite{bavikadi2024geospatial}.  That work is designed to produce faux movement trajectories and not identify actual future regions.  We note that it relies on a different approach (the use of A*) suitable to create a movement trajectory that meets constraints but doesn't provide regions that allow for multiple future trajectories. In contrast, this work examines generating regions with a top-$k$ entailment query necessitated by the nature of the problem and data at hand.

\vspace{-5pt}

\section{Approach}

\subsection{Logic for Maritime Agents}
\label{subsec:prelims}
\noindent\textbf{Logical Language.} To define various aspects of the maritime domain environment, we use an annotated language \cite{ks92,ssTAI22} with temporal semantics \cite{shakAamas13,aditya2023pyreason,bavikadi2024geospatial}. The language is defined with a set of constants that is partitioned into multiple domains ($\domainSet_i \subset \constantSet$), one such subset, $\domainSet_{loc}$, is a set of all potential locations of the vessel in a continuous space (``area of interest'' or AOI) of dimensions $M \times N$.  As usual in first-order logic, we define a corresponding set of variables ($\variableSet$) and a set of predicate symbols ($\predicateSet$). Additional sets of constants include a set $\domainSet_{agt}$ - a set of agents (in our application, maritime vessels) and $\domainSet_{\regionconst}$ - a set of all regions within the AOI (in practice, we compute this based on historical trends ahead of time). When it is relevant, we shall subscript such constants with the upper-right and lower-left locations - e.g. $\regionconst_{l1,l2} \in \domainSet_{\regionconst}$ is a region with upper-right corner $l1$ and lower-left corner $l2$ ($l1,l2 \in \domainSet_{loc}$).  We treat $\regionconst$ as a set of all locations enclosed by the region.
In addition to the first-order logic syntax and semantics, we allow for annotation $[\ell,u]$ (that are elements of a lower semi-lattice structure  $\semiLattice$) which is simply a subset of the unit interval $[0,1]$ - which generalizes both fuzzy and classical logic.  We write an annotated literal $a:[\ell,u]$ to mean that the literal $a$ has truth value associated with interval $[\ell,u]$. We refer the reader to \cite{ks92,ssTAI22} for lattice-theory justification of this approach.  We learned our logic programs in a way to treat these bounds as confidence (see Section~\ref{sec:rule_learning}). 
 We follow the extension of temporal syntax and semantics~\cite{aditya2023pyreason,bavikadi2024geospatial} to form temporally annotated facts (TAFs) and annotated formulae. For an annotated literal $f$ that is true at time $t$, $f_t$ is a TAF. Annotated formulae are constructs formed with operators like $\after(f,f')$. For annotated literals $f,f'$, $\after(f,f')$ is interpreted as $f$ occurs after $f'$.

\begin{example}[Language]
\label{ex:syn}
\vspace{-4pt}
\textit{In our use-case, we consider an agent $agt \in \domainSet_{agt}$ that travels among $loc1, loc2.. \in \domainSet_{loc}$ in an AOI. The agent can be at a location covered by a region $\regionconst \in \domainSet_{\regionconst}$ where $\regionconst \subseteq \domainSet_{loc}$.  
We define domain-specific binary predicate, $\atPred$, where $\atPred(agt, \regionconst)$ is a ground atom for an agent, $agt $, at a location  in $\regionconst$ indicating that the agent is within the region of $\regionconst$.
We also define domain-specific unary predicates formed with $\domainSet_{agt}$ constants: $\nearport$, $\changedirection$, $\highspeed$, $\lowspeed$,  $\hotspot$, $\draught$, $\aisoff$ and $\stay$   (expressing that the agent is near a port, changed its course sharply, has a high/low speed compared to an average, at a high-density hotspot, varied its draught, stopped transmitting AIS signals, and is at an anchor point by staying put for a long duration).}
\vspace{-4pt}
\end{example}

As per previous work on temporal annotated logic~\cite{shakAamas13,aditya2023pyreason,bavikadi2024geospatial}, given the set of all ground literals $\groundLiteralSet$, a set of timepoints $T$, an interpretation $\interpretation$ is any mapping $\groundLiteralSet \times T \to \semiLattice$. We define a satisfaction relationship ``$\models$'' and rules for temporally annotated extensions~\cite{shakAamas13, aditya2023pyreason}.  A program $\Pi$ is a set of TAFs and rules, where each has an annotated atom in the head and a conjunction of annotated formulae in the body.  An interpretation $\interpretation$ is said to satisfy $\Pi$, if and only if $\interpretation$ satisfies every rule and TAF in $\Pi$.  The minimal model is an interpretation that can be thought of everything that can be concluded from deductive inference and commonly used for entailment queries in annotated logic~\cite{ks92,shakAamas13,ssTAI22,aditya2023pyreason} often computed using a fixpoint operator. 
In this work, we slightly abuse the notation of \cite{ks92} and use $\Gamma^*(\Pi)$ to denote the minimal model of $\Pi$.

\noindent\textbf{Initial and Predicted Locations.}  In our problem, we must represent the initial conditions of the agent - in other words, the areas the shipping vessel has traveled in the first part of its voyage before going dark.  We represent this simply with the logic program consisting of a set of TAFs formed with the predicate $\atPred$ introduced in Example~\ref{ex:syn}.  Here, we would expect fine-grain information on the location of the shipping vessel from information such as AIS - to extract each region (the second argument associated with the $\atPred$-formed TAF).  We can think of such an initial logic program, $\Pi_{init}$ being complemented by an additional logic program - also created with TAFs - used to represent the predicted agent's behavior in the future - $\Pi_{pred}$.  Intuitively, the elements of $\Pi_{pred}$ would resemble the elements of $\Pi_{init}$ except that they would occur after the facts of $\Pi_{init}$.  Further, in practice, we would expect regions associated with $\Pi_{pred}$ to be larger than $\Pi_{init}$.  We shall refer to these logic programs $\Pi_{pred},\Pi_{init}$ as \textit{region set program} and provide Example~\ref{ex:regionsets} of this.

\begin{figure}[t]
  \centering
  \includegraphics[scale=.35]{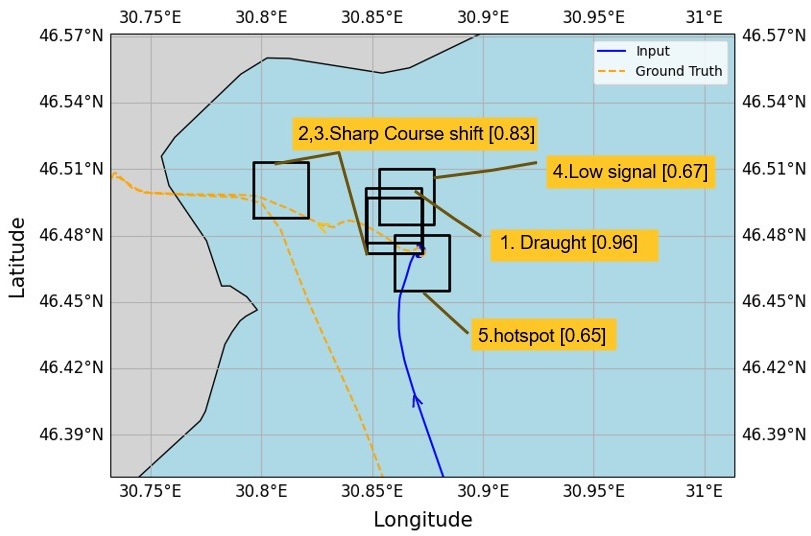}
    \caption{Abduction model predictions. The solid line is the input test sample. The dashed line is the ground truth. Black regions are the generated regions along with confidence and region types.}
    \vspace{-8pt}
    \label{fig:xai}
\end{figure}

\begin{example}
\label{ex:regionsets}
\vspace{-4pt}
 Consider an agent $agt \in \domainSet_{agt}$ in the Figure~\ref{fig:xai} that travels from time $t_1$ to $t_i$-denoted by the solid blue line (we notate timestamps to be a set of timepoints $=\{t_1,.,t_i,..,t_j,..,t_n\}$ with a precedence relationship) and then goes dark after $t_i$-denoted by the dashed line, a time horizon $t_{j}$, then the initial conditions are represented as follows, 
$\program_{init} = \{ $
$ \atPred(agt,\regionconst_{(31.14,46.12),(31.11,46.09)})_{t_1}$\\$,.., \atPred(agt,\regionconst_{(30.88,46.48),(30.86,46.45)})_{t_i} \}$  and the predictions are represented as follows,  $\program_{pred} = \{ $$ \atPred(agt, \regionconst_{(30.87,46.51), (30.85,46.48)})_{t_{j}}$
,\\$ \atPred(agt, \regionconst_{(30.82,46.51),(30.79,46.48)})_{t_{j}}$
,$\atPred(agt, \regionconst_{(30.88,46.48),(30.85,46.45)})_{t_{j}}$
,$ \atPred(agt, \regionconst_{(30.87,46.50),(30.84,46.47)})_{t_{j}}$
,$ \atPred(agt, \regionconst_{(30.87,46.49),(30.84,46.47)})_{t_{j}} \}$-denoted by the 5 corresponding black regions in Figure~\ref{fig:xai}.
\vspace{-3pt}
\end{example}

\noindent\textbf{Behavior Rules.}  We also envision a logic program of a set of behavior rules of what the shipping vessel normally does ($\Pi_{behav}$) with example rules mined from data in Table~\ref{tab:examplerlsRules}.  While it is possible to make these rules function as hard constraints, we instead make them soft constraints and measure how well an agent complies with these rules - enabling us to easily build a parsimony function.

\noindent\textbf{Ground Truth Trajectories.}  Based on historical data, we assume we have trajectory data for a given agent that occurs outside of $\Pi_{init}$.  For a given agent, such a trajectory is simply a series of location-time tuples that were observed in the ground-truth data.  So for agent $agt$, trajectory $ \tau_{agt}^{AIS} = \langle (loc_1,t_1),$ $\ldots,(loc_i,t_i),\ldots,(loc_n,t_n)\rangle$, the ground truth trajectory is $\tau_{agt}= \langle (loc_i,t_i),\ldots,(loc_n,t_n)\rangle$.  We define the notion of entailment of a trajectory at the syntactic level (though it is trivial to derive a semantic version) and provide Example~\ref{ex:entail} of this.  We say the program $\Pi$ entails an agent's trajectory $\tau_{agt}$ if for all $(loc,t) \in \tau_{agt}$ there is some TAF $\atPred(agt,\regionconst)_t \in \Pi$ (which occurs at the same time) such that $loc \in \regionconst$.

\begin{example}
\label{ex:entail}
\vspace{-4pt}
Following the notion built in Example~\ref{ex:regionsets}, the trajectory for agent $agt$ is, $\tau_{agt}^{AIS} = \langle ((31.11,46.00), t_1),..,((30.87,46.47)$\\$,t_i),((30.85,46.48),t_{i+1}),((30.81,46.49),t_{i+2}),..,((31.07,46.00), t_n) \rangle$, then $\program_{init} \bigcup \program_{pred} \models \tau_{agt}^{AIS}$. Note that tuples of $\tau_{agt}^{AIS}$, like $\tau^1,.., \tau^i$ are entailed by TAFs in $\program_{init}$ - $(31.11,46.00) \in \regionconst_{(31.14,46.12),(31.11,46.09)}$, \\..,$(30.87,46.47) \in \regionconst_{(30.88,46.48),(30.86,46.45)}$ and the others can be entailed from $\program_{pred}$ - $(30.85,46.48) \in \regionconst_{(30.87,46.50),(30.84,46.47)}$, and $(30.81,46.49) \in \regionconst_{(30.82,46.51),(30.79,46.48)}$- denoted by black regions 2 and 3 in Figure~\ref{fig:xai}, for instance. Similarly, $\program_{pred} \models \tau_{agt}$.
\end{example}

\subsection{Abducing Agent Trajectories}
\label{sec:abduc}
For a single agent, we can think of finding $\Pi_{pred}$ as an abduction problem.  In other words, given an agent $agt$, initial conditions $\Pi_{init}$, behavioral rules $\Pi_{behav}$, and ground-truth trajectory $\tau_{agt}$ we want to find $\Pi_{pred}$ such that:
\begin{enumerate}
    \item $\Pi_{init}\cup \Pi_{behav} \cup \Pi_{pred}$ is consistent (i.e., $\Gamma^*(\Pi_{init}\cup \Pi_{behav} \cup \Pi_{pred})$ exists).
    \item For each $\Pi_{pred}$ entails $\tau_{agt}$
\end{enumerate}
If these criteria are met, we say $\Pi_{pred}$ is an explanation for\\ $\langle agt, \Pi_{init}, \Pi_{behav}, \tau_{agt} \rangle$.  In this paper, our goal is to find a function that, based on historical data, can return an explanation.  We define an \textit{explanation function} as follows.

\vspace{-3pt}
\begin{definition}[Trajectory Explanation Function]
\label{traj_abd}
Given agents $agt^1,\ldots,agt^n$, initial condition programs $\Pi^1_{init},\ldots,\Pi^n_{init}$, behavioral rules $\Pi_{behav}$, and trajectories $\tau_{1}^{AIS},\ldots,\tau_{n}^{AIS}$, we say an explanation function $f_{E}$ that takes as arguments an agent and two programs and returns a region set program such that $f_{E}(agt^i,\Pi^i_{init},\Pi_{behav})$ is an explanation for  $\langle agt^i, \Pi^i_{init}, \Pi_{behav}, \tau_{i}^{AIS} \rangle$.
\end{definition}
\vspace{-2pt}

We note that Definition~\ref{traj_abd} is quite strict as it requires the result of $f_{E}$ to produce a region set that models the entire trajectory for all agents.  At the same time, it does not distinguish among different explanations.  We introduce an approximation, $\hat{f_{E}}$ that is designed to meet the entailment requirement for as many agents as possible.  Our solution is to leverage a notion of parsimony, defining $\hat{f_{E}}$ in terms of a parsimony function ($\parsimony$) - which maps agents and logic programs to scalars.  The idea is to use $\parsimony$ to measure the quality of an explanation so that we can find quality explanations that cover most of the ground truth trajectories.  We provide the following examples of such a function.
\vspace{-7pt}

\begin{eqnarray*}
    \hat{f_{1}}(agt,\Pi_{init},\Pi_{behav}) = \arg \max_{\Pi'} \parsimony(agt,\Pi_{init}\cup \Pi_{behav} \cup \Pi')\\
    \hat{f_{2}}(agt,\Pi_{init},\Pi_{behav}) = \{ \arg \max_{\phi} \parsimony(agt,\Pi_{init}\cup \Pi_{behav} \cup \{\phi\})\}
\end{eqnarray*}
\vspace{-5pt}

In these two examples, we note the first has a combinatorial flavor - finding the best set of regions, while the second identifies the best singleton set - a notion that we can extend to find the top $k$ singletons (which correspond to the top $k$ regions formed with the $\atPred$).  This can be easily solved by multiple entailment problems for each relevant singleton formed from atoms created with set $\domainSet_{\regionconst}$ (which we assume is known a-priori).  We also note that the computation of $\hat{f_2}$ can be computed in linear time (in the number of TAFs) which results directly from the prior results on annotated logic~\cite{ks92,ssTAI22} and allows us to leverage existing efficient implementations~\cite{aditya2023pyreason}.  We verify this empirically (Figure~\ref{fig:runtime}).  In this work, we examine the top-$k$ variant of $\hat{f_{2}}$ and provide empirical evidence that supports it.
In practice, we compute top-k regions - corresponding to the TAF $at(agt,\regionconst)$ (picking $\regionconst$ from $\domainSet_\regionconst$) in parallel.

\subsection{Rule-Based Agent Behavioral Modeling}
\label{sec:rule_learning}
As described in Section~\ref{subsec:prelims} we assume that there exists a set of rules $\Pi_{behav}$ specifying the behavior of the agents.  While we could design $\Pi_{behav}$ to allow for hard constraints on consistency (and while there are good reasons for doing so), we instead leverage the fuzzy nature of our underlying logic (as described in Section~\ref{subsec:prelims}) which can then allow us to easily build an explainable parsimony function $\sigma$.  Again, this function takes an agent and a logic program as arguments (and the logic program, $\Pi$, is the union of the initial conditions $\Pi_{init}$ and behavior rules $\Pi_{behav}$) and returns a scalar.  As we use the logical paradigm of \cite{ks92,ssTAI22}, each logical atom is associated with a subset of the unit interval - $[\ell,u]$.  In this work, define the parsimony function as the aggregate over the lower bound of the interval, formally:
\vspace{-3pt}
\begin{eqnarray*}
\parsimony_t(agt,\Pi) & =& lb\Bigl ( \Gamma^*(\Pi )(normal(agt))(t) \Bigr)
\end{eqnarray*}

Intuitively, we have a predicate $normal$, such that atoms formed with that predicate are annotated with an interval measuring the agent's level of normalcy.  The minimal model of the program, $\Gamma^*(\Pi )$ provides this annotation for a particular atom - here $normal(agt)$ (the normalcy of agent $agt$) and time $t$ (we can define $\sigma$ for a particular time - in practice we use the maximum time as it allows us to cover long-term predictions).  Finally, $lb$ returns the lower bound of the interval (as we will learn rules in a manner where we set the upper bound to $1$ to easily ensure consistency).

\noindent\textbf{Rule Learning Algorithm.}  From the training set, a set of rules is learned to model the normal behavior of the vessels based on the historical co-occurrences of periodic sequences among similar types of ships in similar waters. They are learned in a method akin to rule learning in~\cite{APTL, bavikadi2024geospatial} where we restrict the body to have a single sequence of movement, refer Algorithm~\ref{alg:rulelearning}. These rules are population-specific among the vessels. Here, consider $\tau^r$ to be a set of the associated region of the trajectory.
We note that Algorithm~\ref{alg:rulelearning} is quite efficient. It scans all trajectories in a given data. The quantity of trajectory size in terms of regions can be treated as a constant as it's from a data source. Hence, it turns out that Algorithm~\ref{alg:rulelearning} is linear in terms of the size of the dataset (number of trajectories). 

\begin{algorithm}
    \caption{ Behavioral Rule Learner}
    \label{alg:rulelearning}
\begin{algorithmic}[1]
\State \textbf{Input:} A set of trajectories $\trajectoryset$, atom $normal(agt)$
\State \textbf{Output:} A set of rules $\program$

\Function{Rules}{$ Body$}
\State $\program$ $\gets \emptyset$

\ForAll{$ moves \in Body$ }
    \If{length($moves$) = 2}

        \State $\textit{mov} \gets Body[moves][0]$
        \State $\program \gets \program \bigcup$ $\{ normal(agt): [ \frac{Body[moves]}{Body[mov]}, 1] \gets \bigwedge_{m \in moves} m(agt) \}$
     \EndIf
\EndFor
\State \textbf{return} $\program$
\EndFunction
\Function{TrainModel}{$\trajectoryset$}
    \State Initialize dictionary $Body \gets \emptyset$
    \ForAll{$\trajectory^r$ in $\trajectoryset$}
        \For{$n \gets 1$ \textbf{to} length($\trajectory^r$)-1}
                \State $Body[\trajectory^r[n]] \gets Body[\trajectory^r[n]]+1$ 
                \State $Body[\trajectory^r[n-1]] \gets Body[\trajectory^r[n-1]]+1$ 
                \State $Body[(\trajectory^r[n - 1], \trajectory^r[n ])] \gets Body[(\trajectory^r[n - 1],\trajectory^r[n])]+1$ 
            
                \EndFor
            \EndFor
    
    \State $\program \gets $Rules$(Body)$
    \State \textbf{return} $\program$
\EndFunction

\end{algorithmic}
\end{algorithm}

\begin{table*}[t!]

\caption{Example Rules Mined From Historical Data}
\label{tab:examplerlsRules}
\begin{tabular}{@{}p{0.58\textwidth}p{0.38\textwidth}}
\toprule
Rule & Natural Language\\

\midrule

$normal(AGT): [0.8,1] \leftarrow{} nearport(AGT):[1,1] \wedge high-hotspot(AGT):[1,1] \wedge \after(high-hotspot(AGT), nearport(AGT)): [1,1] $
 &

\textit{Example Multi-hop rule.} The confidence of a vessel exhibiting normal behavior is at least $0.8$ when the agent goes from a near port to a high-hotspot region in more than a single movement.\\
\vspace{0.2pt}
$normal(AGT): [0.9,1] \leftarrow{} low-speed(AGT):[1,1] \wedge sharp-course-change(AGT):[1,1] \wedge \after(sharp-course-change(AGT), low-speed(AGT)): [1,1]$ & \vspace{0.2pt} \textit{Example Single-hop rule.} The confidence of a vessel exhibiting normal behavior is at least $0.9$ when the agent changes its course direction after lowering its speed in a single movement.\\
\bottomrule
\end{tabular}
\end{table*} 

\noindent Here, the movement is considered to be among regions representing features like port regions, density-based historical hotspots, anchor points, destinations, and typically observed maritime features (speed over ground, course over ground, and heading). The observed maritime features include the regions where we historically observe the feature’s spikes in their usual values, for instance, vessels exhibiting high speeds in specific regions. We define two kinds of rules based on the movement from the current region. It could be one (single-hop rules (SH)) or multiple hops (multi-hop rules (MH)) away to the next region. The intuition is to capture movements that occur eventually and in the next movement from the current region. 
Sample rules that we mined from maritime vessel data are shown in Table~\ref{tab:examplerlsRules}. The annotations on the head of the rules note the measure of confidence in the normalcy of the rule.


\section{Experimental Results}
\label{sec:exp}
\noindent\textbf{Setup.} We parsed  Automatic Identification System (AIS) data of $614$ vessels across the Black Sea AOI from January 2022 to March 2023. This involves the trajectory data $\tau$ of each vessel in addition to its dynamic and statistical information. This data has trajectories of the length $2$ to $165,000$ data points (i.e., the vessel's latitude, longitude, timestamp, other features~\cite{ais}) that span from $1$ to $264$ days.  For all our experiments, we use a high memory compute node, Dell PowerEdge R6525 with the AMD EPYC 7713 64-Core Processors and 2TB RAM, along with three A30 GPUs. The region size is fixed arbitrarily at $0.025\degree \times 0.025\degree$ which comes to $5.45 km^2$ in our AOI for our experiments unless specified. 
Extending prior work~\cite{groupwaterarea, groupshipwdestination} where similar vessels were grouped, we perform trajectory clustering~\cite{predwithcluster} with DBSCAN to group trajectories into $9$ subset and we report average metrics across all clusters for both our method and our deep learning baseline.
As the ground truth data for dark activity has limited availability, and we aim to generate regions at a future time, we mask each test trajectory to obtain a partial trajectory along with the ground truth - adapting from the setup in the prior work~\cite{dlbaseline,dstnet,txlstm, tf, tp}. This strategy involves using historical data due to the scarcity of external ground truth dark activity, resulting in generating regions informed by historical behavior. Furthermore, our approach is data-efficient, enabling analysts to derive a set of rules even with a limited number of trajectories. Results with available limited ground truth dark activity can be found in the Appendix at \url{https://arxiv.org/abs/2502.01503}.

The masked part is considered the ground truth ($\tau_{agt}$) while the unmasked part is used to set the initial condition ($\Pi_{init}$). We mask half the trajectory from its midpoint in all our experiments unless specified.  
Given $\Pi_{init}$, the model generates top $k$ regions (most relevant regions till a time horizon $t$) to identify the dark vessel.
\vspace{2pt}

\noindent\textbf{Methods.}  We examine three methods, described as follows.

\noindent\textit{Random baseline (RND).}
The random method randomly generates regions from the AOI grid. The AOI grid is formed with cells of the fixed region size. The average performance of three random generators is reported.

\noindent\textit{Deep learning baseline (DL).}
For the DL baseline, we use a sequence-to-sequence model~\cite{dlbaseline} to predict future trajectories. To perform a comparable evaluation, the predicted sequence is mapped to regions in the AOI grid. We also evaluated a deep learning baseline trained on all the data (DL-ALL), which generally was not performant beyond $k=4$ limiting its F1 - we include results from that model only in experiments where it significantly outperforms DL models on subsets.  We experimented with variants of \cite{dlbaseline} with alternative architectures to mimic similar to point-based prediction models~\cite{dstnet,txlstm} but these achieved worse results than DL and DL-ALL.

\noindent\textit{Abduction (ABD).}
The abduction method uses training data to obtain a set of regions (which is the subset of the AOI grid), from which it learns SH rules to obtain $\program$. Given a test trajectory, it then generates top $k$ regions using $\hat{f_{E}}$ via abductive inference.

\noindent\textit{Metrics.} We report precision as the fraction of returned regions that contain points in the ground truth trajectory.  Likewise, recall is the ratio of returned regions containing ground truth points to all regions containing irredundant points from the ground truth trajectory.  The F1 is the harmonic mean of precision and recall.

\vspace{-10pt}
\subsection{Experiments}

We examine the ABD, RND, and DL approaches when applied to AIS data. We first inspect the area efficiency, which has practical significance. We then evaluate the methods for long-term reasoning capabilities. Further, we compare all approaches as a function of $k$ in a standard setting. We also provide hyperparameter sensitivity concerning region size and ablation studies for $\program$ (based on different rule types), and the versatility to masking methods of the test trajectory. Finally, we assess ABD while limiting the training data before concluding with interpretability of results in ABD.

\begin{figure}[t]
    \centering
    \begin{subfigure}[a]{0.49\textwidth}
        \includegraphics[scale=.42]{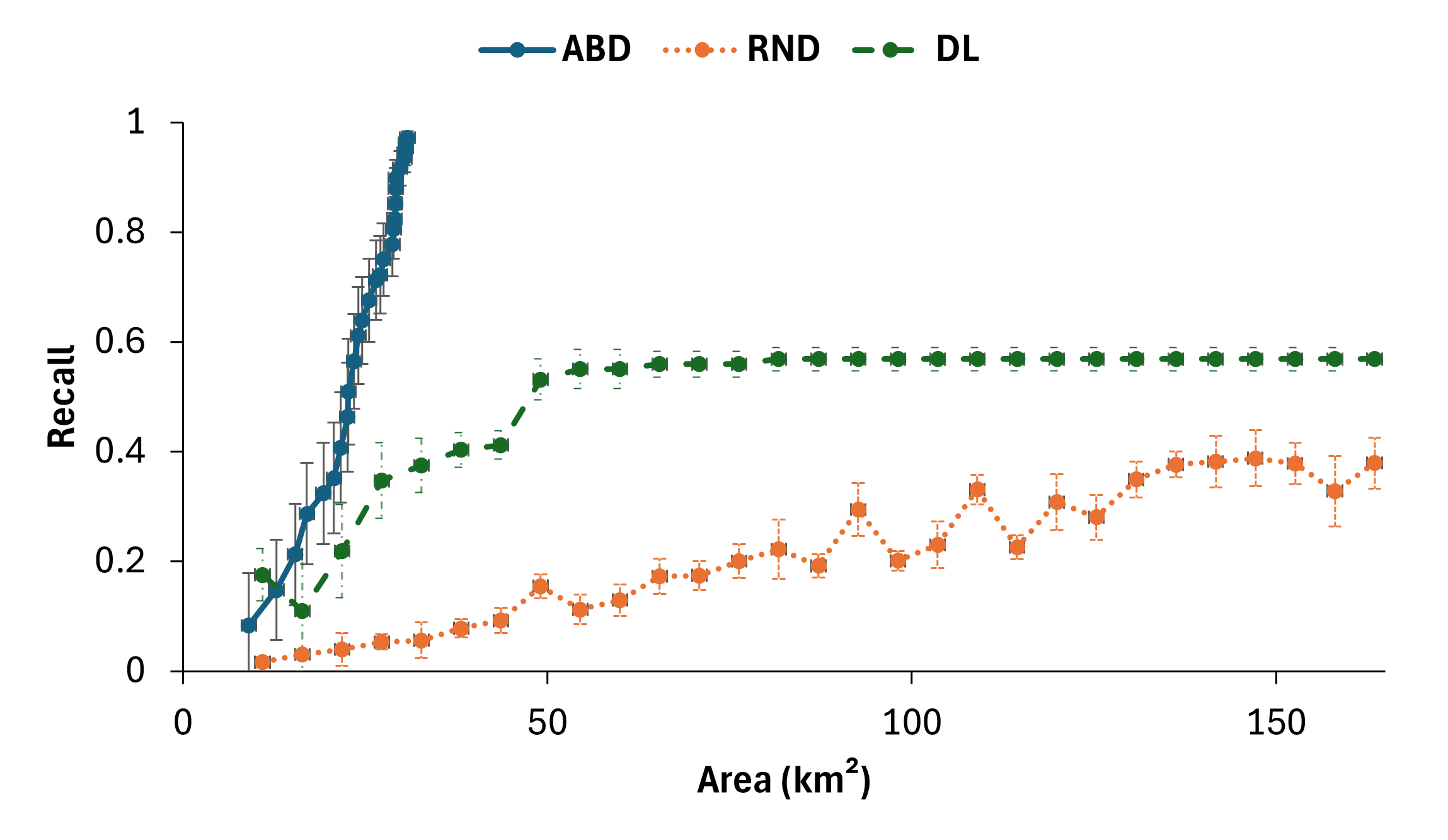}
        \caption{Recall vs. Area}
        \label{fig:recall_area}
    \end{subfigure}
    \hfill
    \begin{subfigure}[b]{0.49\textwidth}
        \includegraphics[scale=.39]{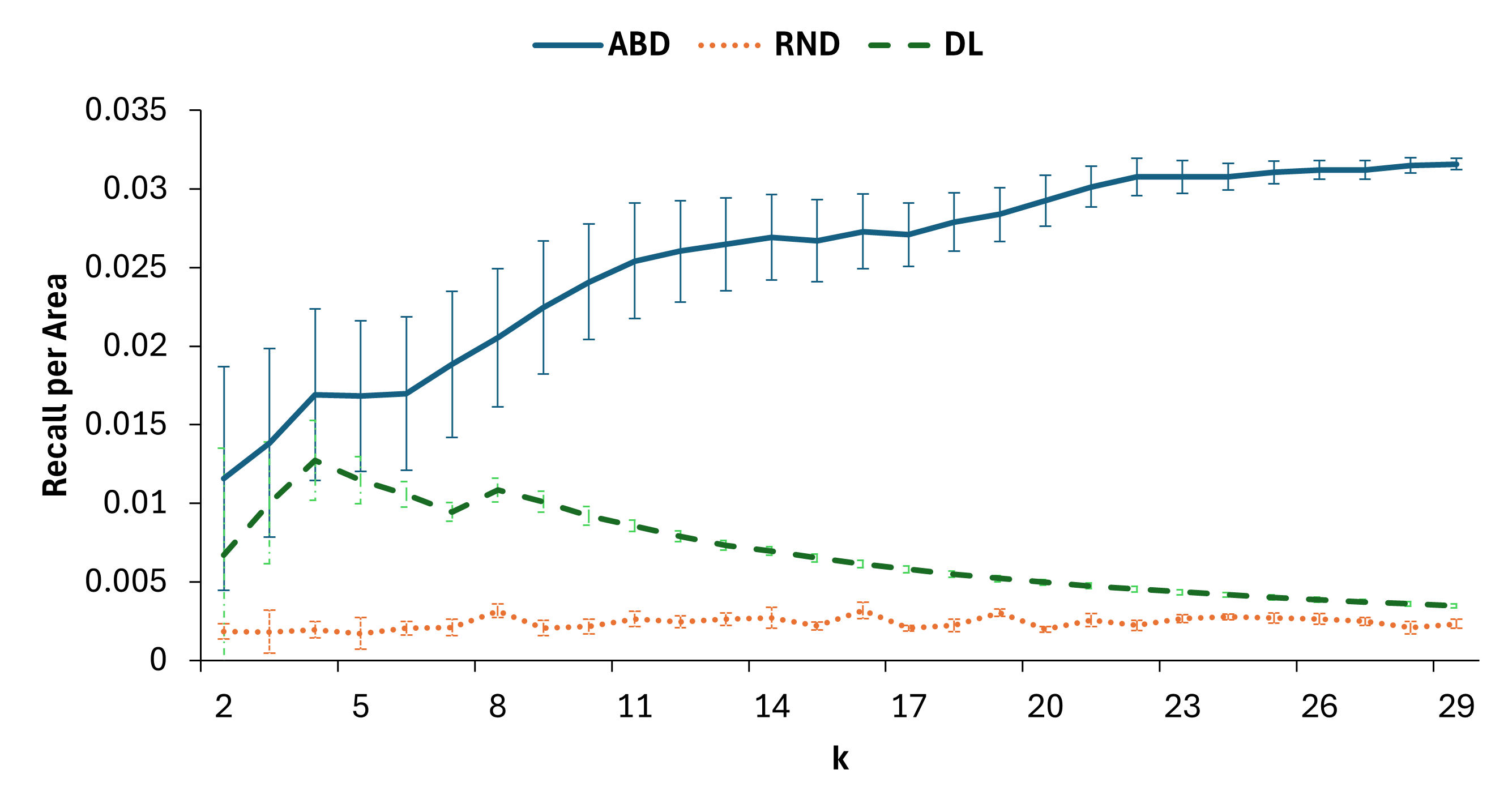}
        \caption{Recall per km$^2$  vs. k}
        \label{fig:recall_per_area_k}
    \end{subfigure}
    \caption{Area Efficiency: (a) Relationship between Recall and Area, (b) Recall per km\textsuperscript{2} as a function of \(k\).}

    \label{fig:area_efficiency}
    \vspace{-10pt}
\end{figure}

\noindent\textbf{Area Efficiency.} In our application, we wish to identify the greatest number of locations for dark vessels while searching the smallest area possible - as identification of dark vessels would require resources such as aerial or satellite imaging.  We examine recall as a function of area in Figure~\ref{fig:area_efficiency}a. We found that recall for ABD saturates at $30$km$^2$ - achieving a recall of $0.99$, which is $157\%$ higher recall than DL for that area.  DL meanwhile saturates at $81.75$km$^2$ - achieving recall of only $0.57$.  This difference suggests that ABD provides more efficiency per unit area.  To further investigate this efficiency, we examine how it trends as a function of $k$ (number of regions) in Figure~\ref{fig:area_efficiency}b.  It turns out that the recall per square kilometer monotonically increases with $k$ for ABD while it decreases for the baselines.  This implies that ABD can continue to produce quality regions. This is significant for practitioners because, when additional search resources are available, ABD can continue to improve search efficiency with the addition of more search resources.

\begin{figure}[t]
  \centering
  \includegraphics[scale=.44]{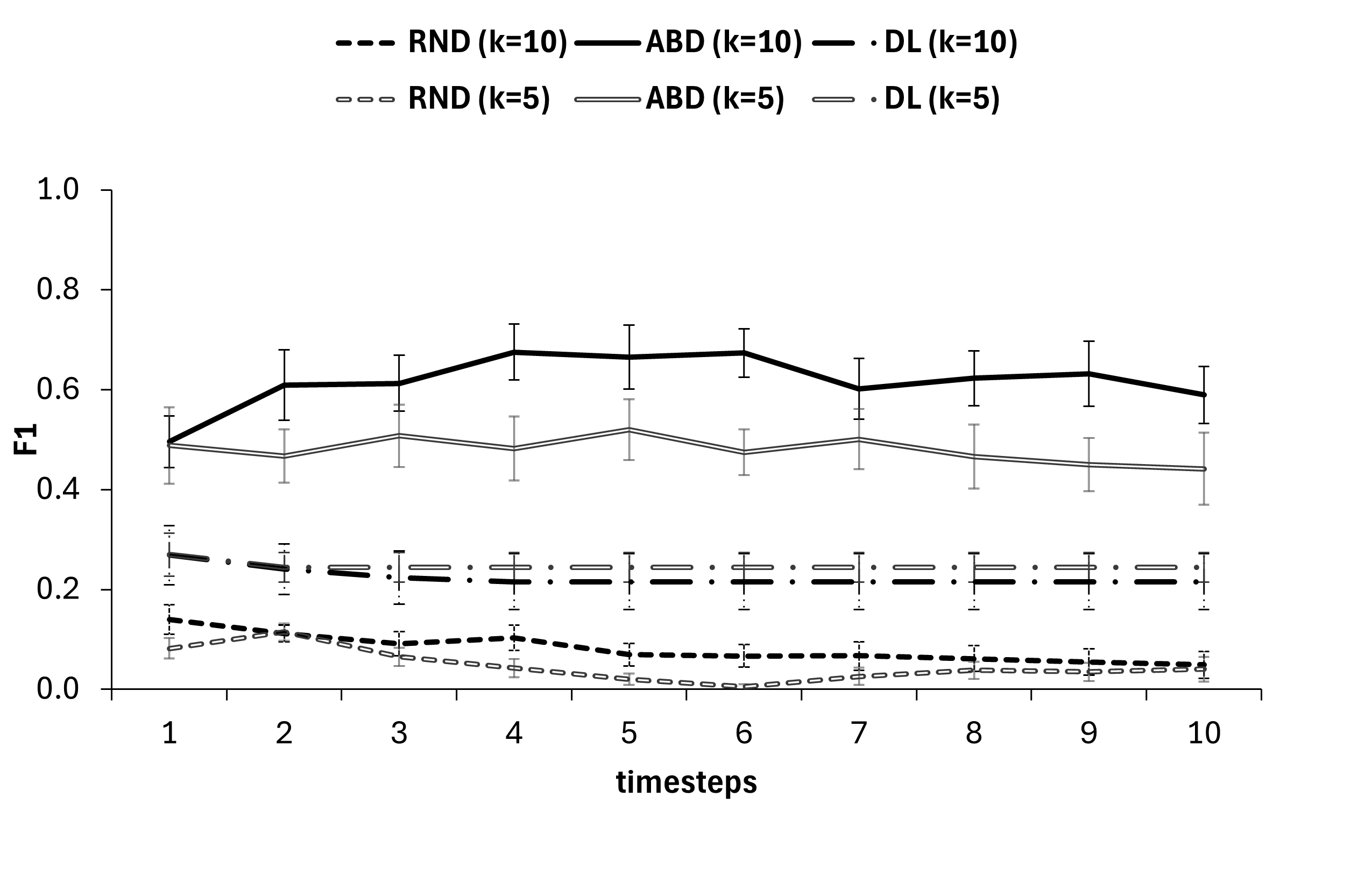}
    \caption{Long-term reasoning. F1@$\{$k=5,k=10$\}$ for ABD, DL, and RND baselines.}
    \label{fig:longterm}
    \vspace{-10pt}
\end{figure}

\noindent\textbf{Long-term Reasoning.} The prior experiments examined performance under the assumption of a fixed time horizon.  Next, we examine performance across multiple time horizons and show the results in Figure~\ref{fig:longterm}.  Here we examine each approach with different settings for $k$ but find that ABD again consistently outperforms other methods in terms of F1.  We also note that ABD is the only approach where an increase in $k$ improves results (e.g., DL achieves poorer performance with $k=10$ vs. $k=5$).  This suggests that our previously described efficiency results likely hold to the case of multiple time units while DL converges by leveling off after the first time horizon. This illustrates that with increasing time horizons, DL is not able to predict long-term trajectories.

\begin{figure}[t]
    \centering
    \begin{subfigure}[a]{0.45\textwidth}
    \centering
        \includegraphics[scale=.46]{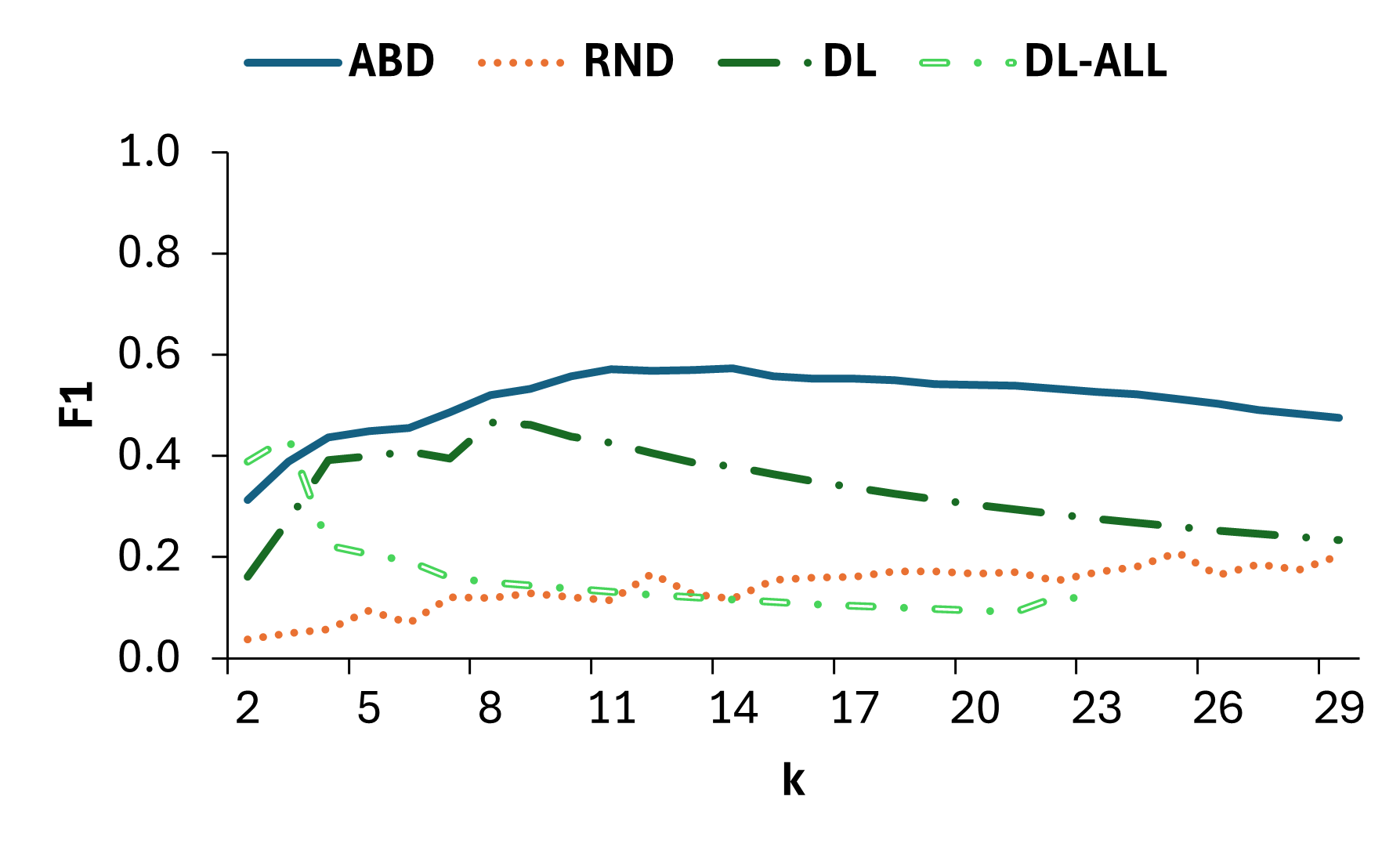}
        \caption{F1@k metric}
        \label{fig:base_f1}
    \end{subfigure}
    \hfill
    \begin{subfigure}[b]{0.45\textwidth}
    \centering
        \includegraphics[scale=.35]{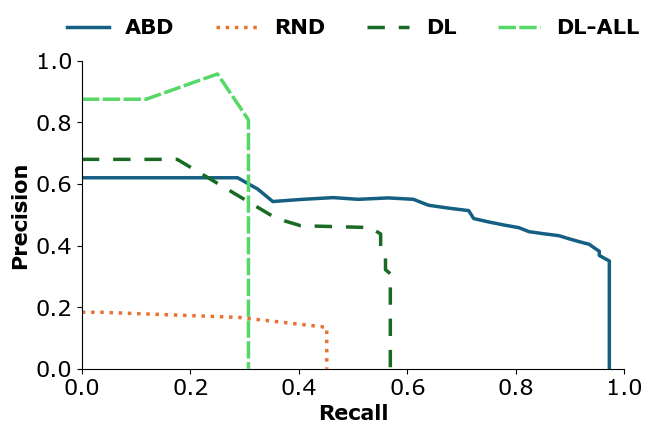}
        \caption{Precision-Recall curve}
        \label{fig:base_pr}
    \end{subfigure}
    \vspace{-2pt}
    \caption{Comparison of (a) F1@k metric and (b) Precision-Recall curve.}
    \label{fig:base}
    \vspace{-10pt}
\end{figure}

\begin{figure}[t]

    \begin{subfigure}[a]{0.49\textwidth}
        \includegraphics[scale=.43]{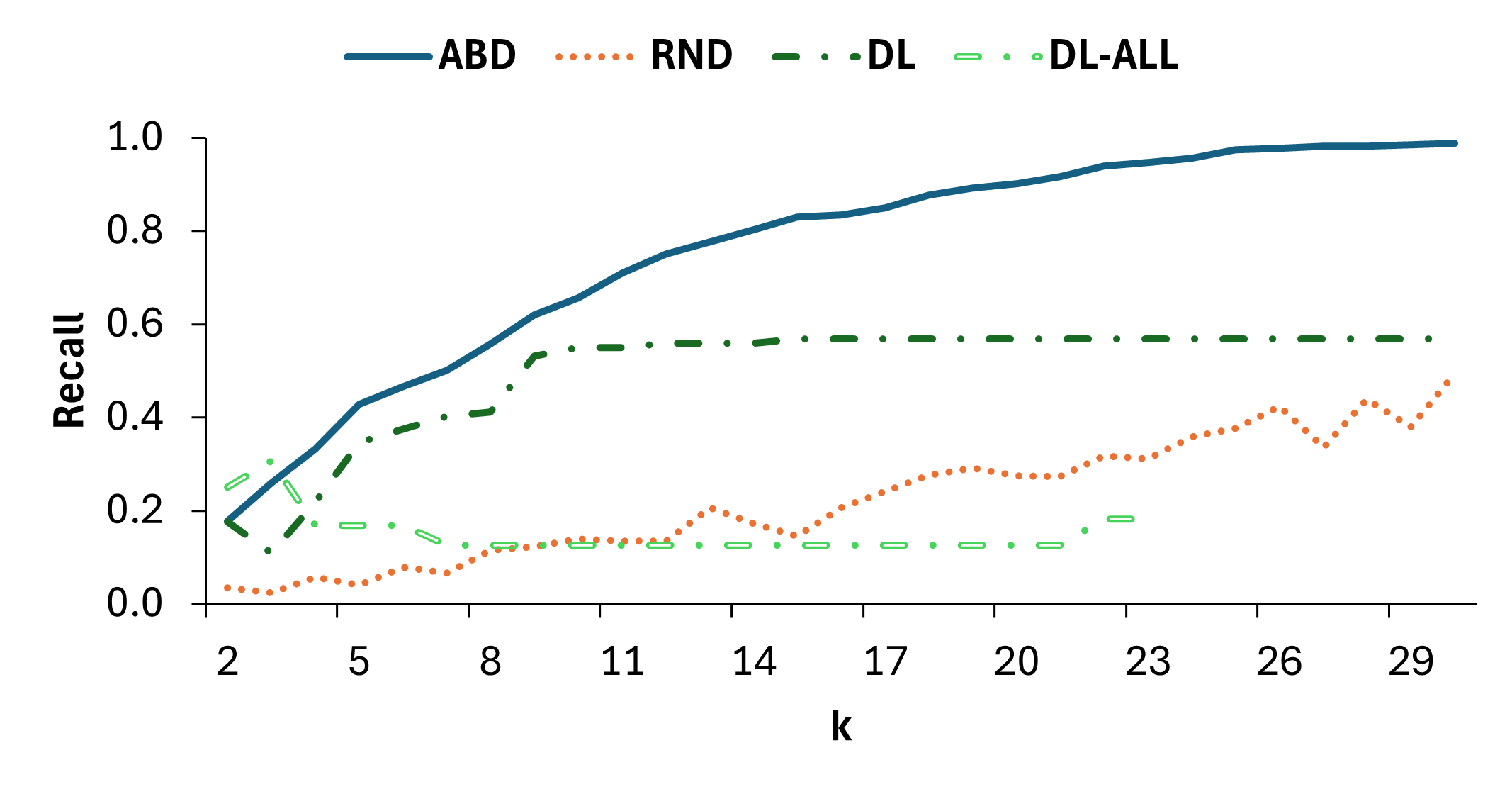}
        \caption{Recall@k metric}
        \label{fig:base_rec}
    \end{subfigure}
    \hfill
    \begin{subfigure}[b]{0.49\textwidth}

        \includegraphics[scale=.46]{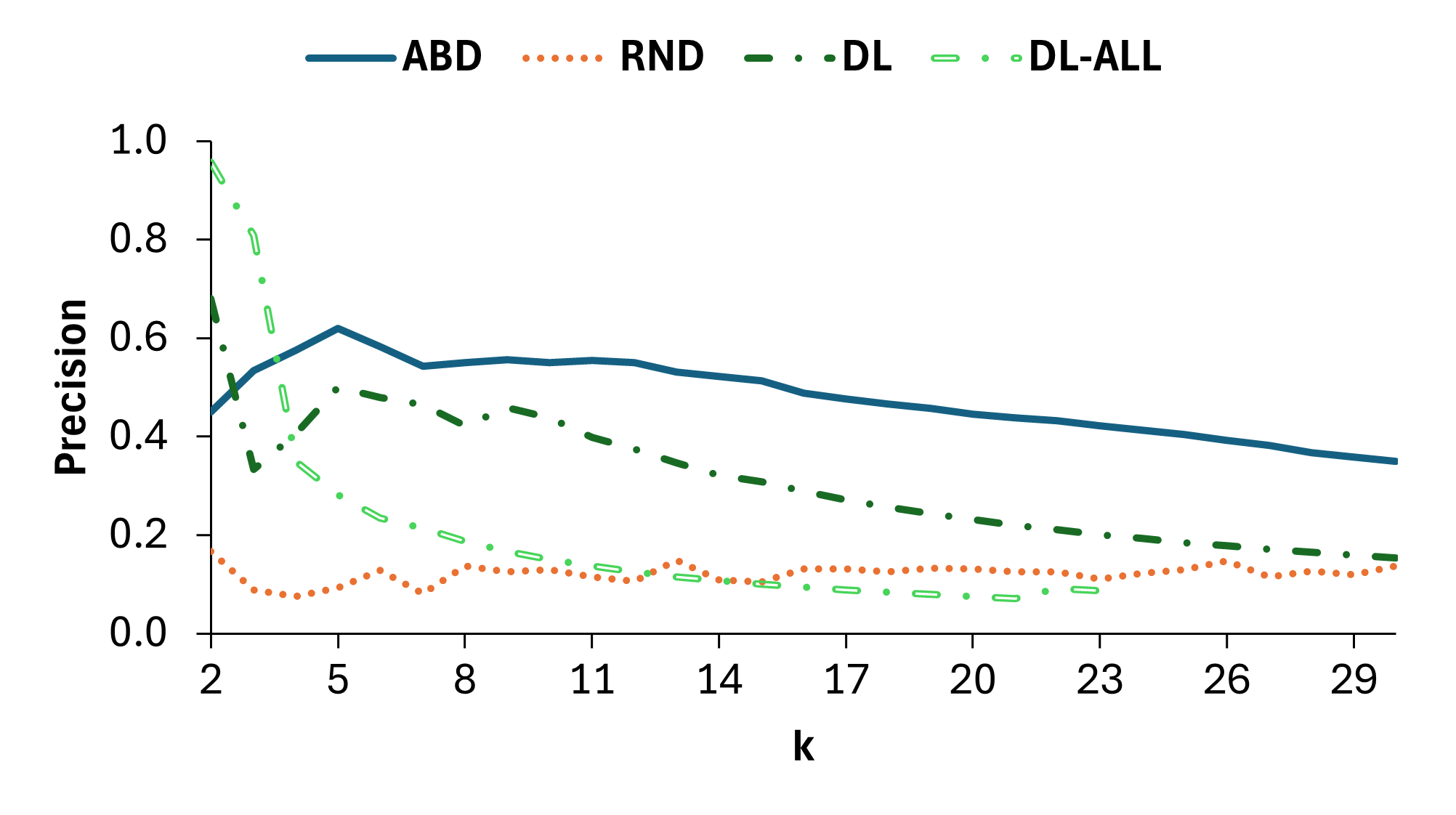}
        \caption{Precision@k metric}
        \label{fig:base_prec}
    \end{subfigure}
    \vspace{-2pt}
    \caption{Comparison of ML metrics-  (a) Recall@k  and (b) Precision@k.}
    \label{fig:base}
    \vspace{-10pt}
\end{figure}

\noindent\textbf{Vessel Recall and Accuracy.} We examine ABD, DL (when trained on the subset datasets individually, as well as on the entire train set), and RND allowing for different values of $k$ (number of regions). Note that the default DL is DL-Subset.  Figure~\ref{fig:base_f1} shows that across all settings of $k$, ABD outperforms all other methods in terms of F1- and ABD on average provides a $51\%$ increase over DL.  For higher values of $k$, DL starts to converge with the random baseline (around $k=28$) while ABD maintains approximately double the F1 score.  When we examine the precision-recall curve in Figure~\ref{fig:base_pr}, we gain an intuition as to why the F1 flags for the DL approach - and the answer is that the recall of DL saturates at $0.57$- indicating limited value in adding more regions (increasing $k$) where ABD can obtain a recall approaching $1$ while increasing $k$, with graceful degradation of precision. ABD has a $476\%$ increase in the recall by adding more regions up to $k=30$.
In this experiment, we also recorded the results of DL-ALL (a single model trained with the whole dataset instead of the sub-datasets). For $k=3,4$, DL-ALL gave the highest F1 due to larger precision values (as seen in Figure~\ref{fig:base_prec}) but the performance degrades by $56\%$ for a unit increase in $k$, and as $k$ increases to $30$, it degrades further. Further, as seen in Figure~\ref{fig:base_pr}, similar to DL, DL-ALL also saturates and does not achieve a recall beyond $0.3$, explaining the decrease in F1. DL-ALL performed $50\%$ lower on F1 when compared to standard DL and the highest F1 for time horizons close to the present.  While this is less relevant for our current application, it may provide insight for further inquiry (e.g., a neurosymbolic approach leveraging both abduction and a model trained with large data).

\begin{figure}[t]
  \centering
  \includegraphics[scale=.44]{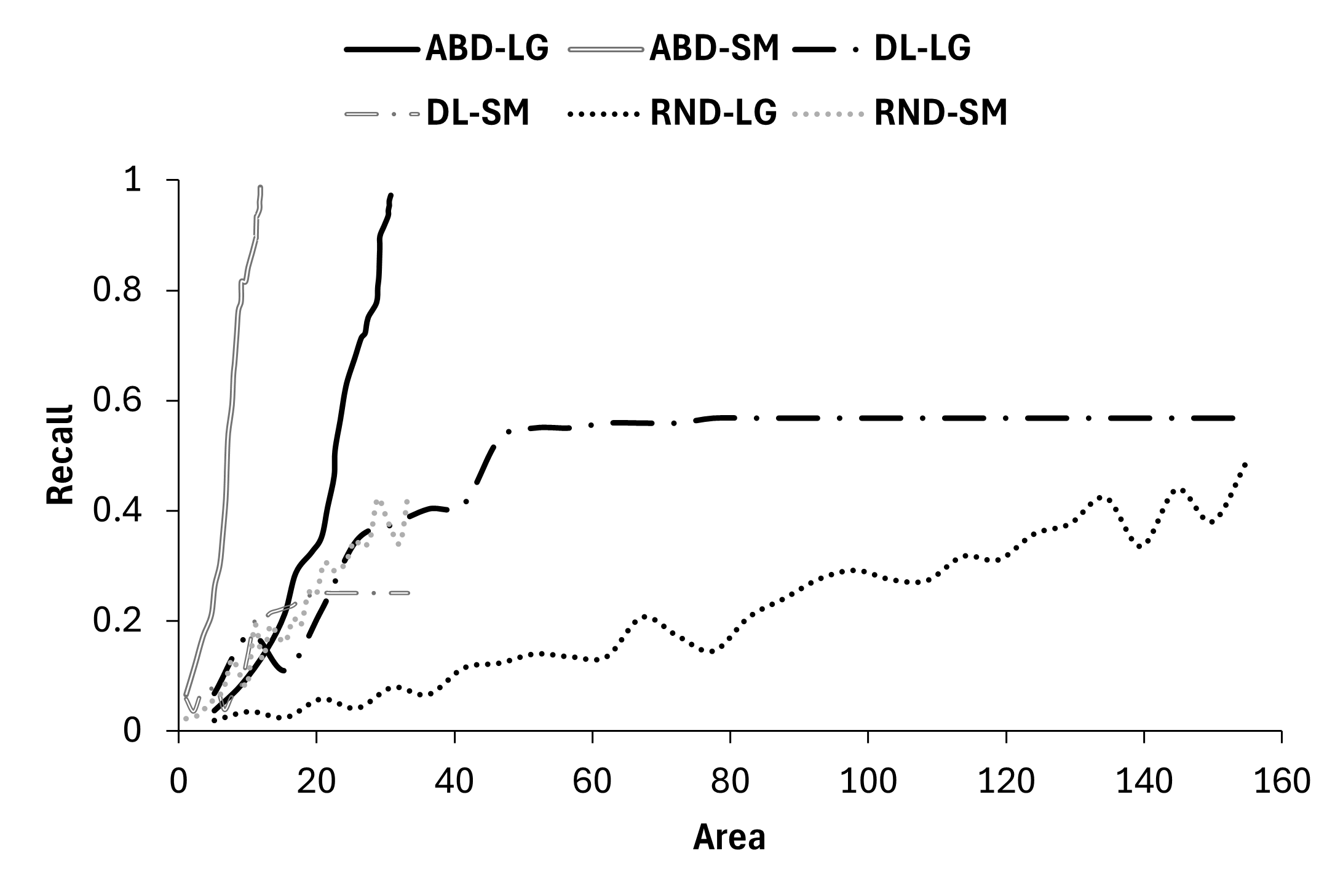}
    \caption{Region size sensitivity for different region-sizes (of LG size = 5.45$ km^2$ and SM size= 1.1 $km^2$). }
    \label{fig:regsize}
    \vspace{-10pt}
\end{figure}

\noindent\textbf{Region Size Sensitivity.}   
In the aforementioned experiments, we determined the region size by considering the computational efficiency of rule learning and generating regions with fair coverage- so that a single region does not end up covering the entirety of the vessel-search space. We now call that setting LG, while the setting SM is when we reduce the region size by $80\%$. 
Note that reducing the region size (SM) is effective by itself as seen by RND-SM achieving comparable performance to DL-LG up to a certain extent. However, ABD outperforms all baselines particularly when the region size is reduced.
The curve is steeper for ABD when the region size is decreased by $80\%$ from LG to SM depicted in Figure~\ref{fig:regsize}, while that of DL resembles its performance in the earlier experiment.  We found that not only that our results maintain with reduced region size, but they also led to improved performance in ABD (reducing the total search area by about 60\%) while the reduction in region size did not meaningfully change the performance of DL.

\begin{figure}[htbp]
    \centering
    \hspace{-0.3cm}
    \begin{subfigure}[b]{0.2\textwidth}
        \centering
        \hspace{-1cm}
        \includegraphics[width=1.11\textwidth]{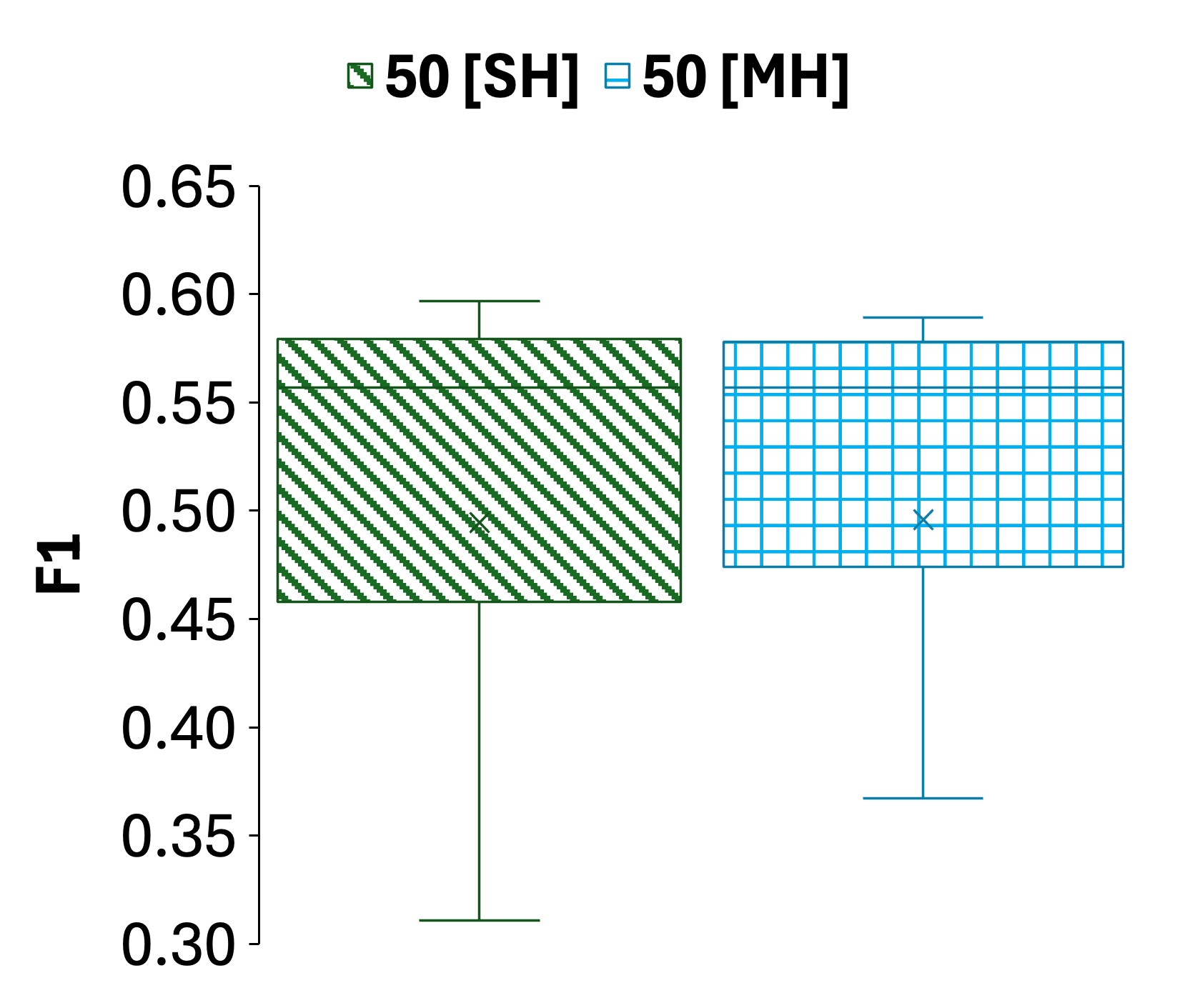}
        \caption{Ablation with Rule type}
        \label{fig:abln_rule}
    \end{subfigure}
    \hspace{-0.65cm} 
    \begin{subfigure}[b]{0.23\textwidth}
        \centering
        \hspace{-0.5cm}
        \includegraphics[width=1.23\textwidth]{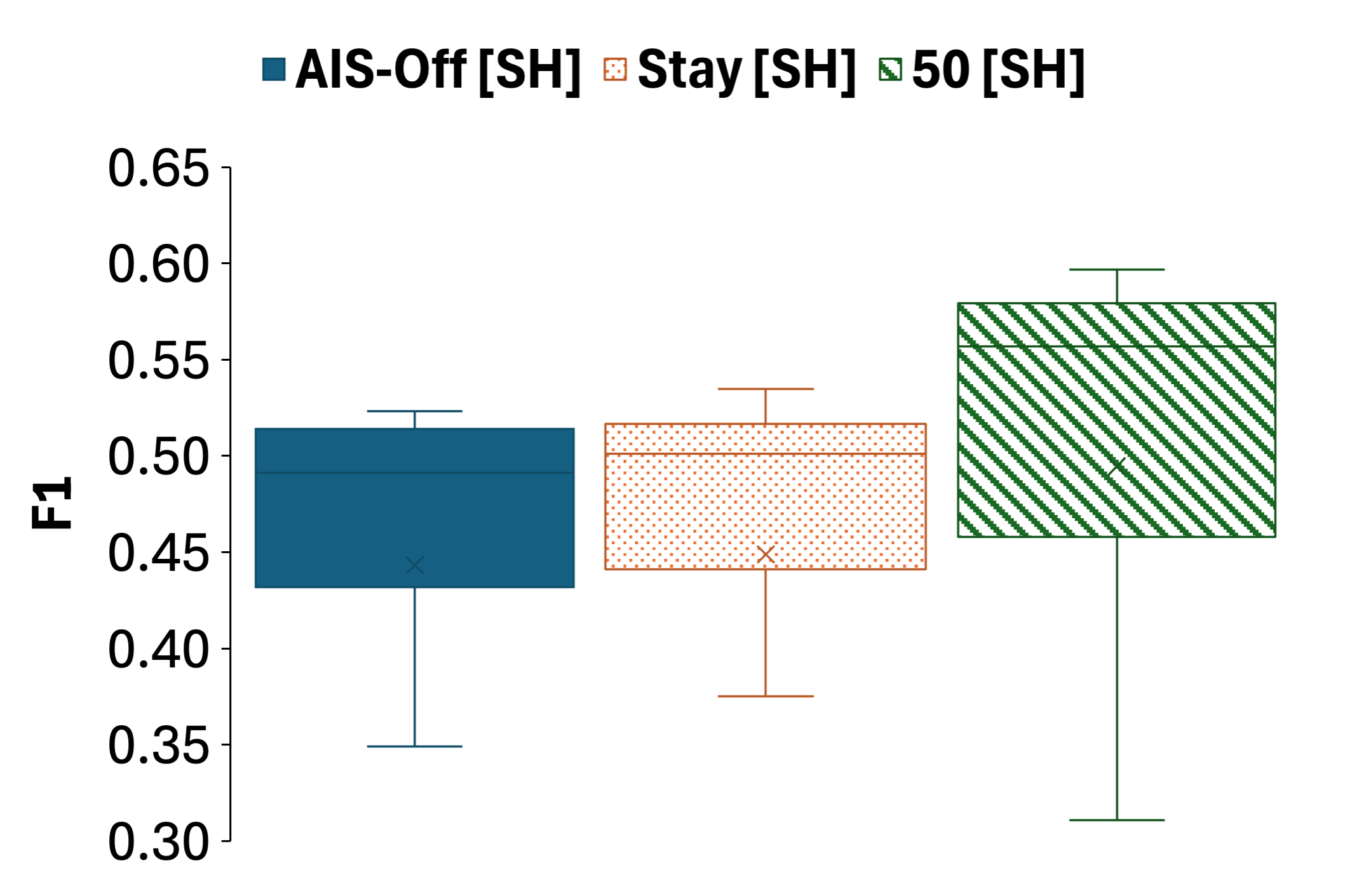}
        \caption{Masking Sensitivity}
        \label{fig:abln_mask}
    \end{subfigure}
    \vspace{-2pt}
    \caption{F1 for various masking methods (AIS-Off, Stay, Baseline 50) and rules (SH, MH).(a) Rule Type  and (b) Masking.}
    \label{fig:main}
    \vspace{-8pt}
\end{figure}

\noindent\textbf{Ablation by Rule Type and Masking Sensitivity.}  As described in Section~\ref{sec:rule_learning} we developed several methods to learn rules (see Table~\ref{tab:examplerlsRules}). In Figure~\ref{fig:abln_rule}, our abduction (ABD) approach still works well for different kinds of rules like single-hop and multi-hop rules. Note that single hop has a slightly wider range of F1 scores with respect to the lower extremes by $0.05$ while the upper extremes and medians are similar.
Additionally, we also wanted to examine the impact of the type of masking on the results - from a practical standpoint to model applications for detecting deceiving vessels who tamper with their AIS transmitter. Here, the masking would start from a point where the AIS is typically tampered with. Different masking strategies include masking the test trajectory at locations when the AIS signal is not received (‘AIS-Off’) or the AIS co-ordinates stays put (‘Stay’) for an unusually long period of time and at the middle of the trajectory (baseline-‘50’). We found that for alternative masking strategies, the median F1 scores decrease by no more than 0.07 from the baseline. This demonstrates the model's adaptability to various use cases, including the detection of dark vessels and identifying vessels that manipulate their AIS to evade sanctions.

\begin{figure}[]
  \centering
  \includegraphics[scale=.44]{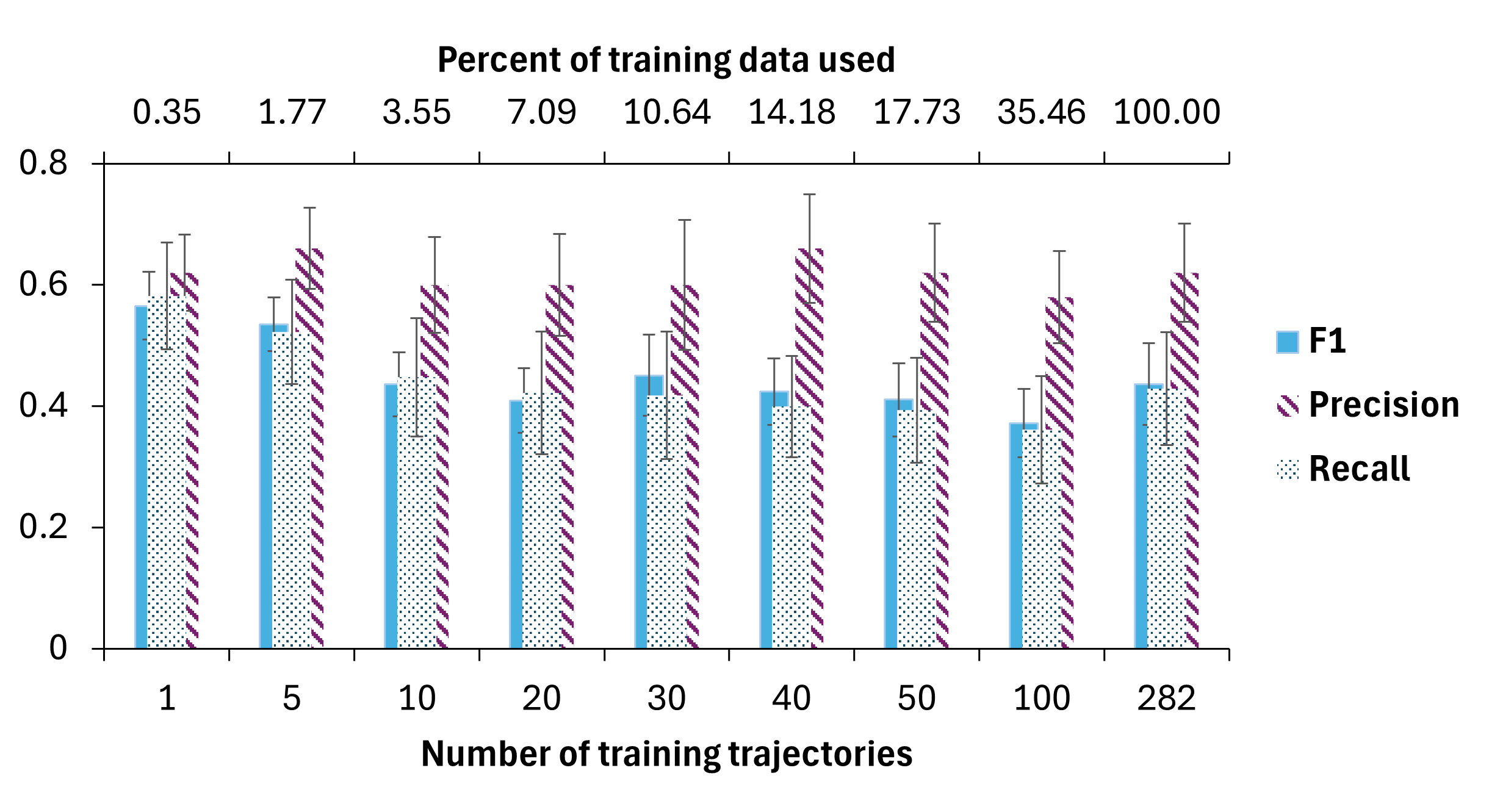}
    \caption{ Evaluation of ABD with limited training data (trajectories). ABD performed smoothly, with variations in the number of training trajectories. }
    \label{fig:eff}
    \vspace{-8pt}
\end{figure}

\begin{figure}[]
  \centering
  \includegraphics[scale=.47]{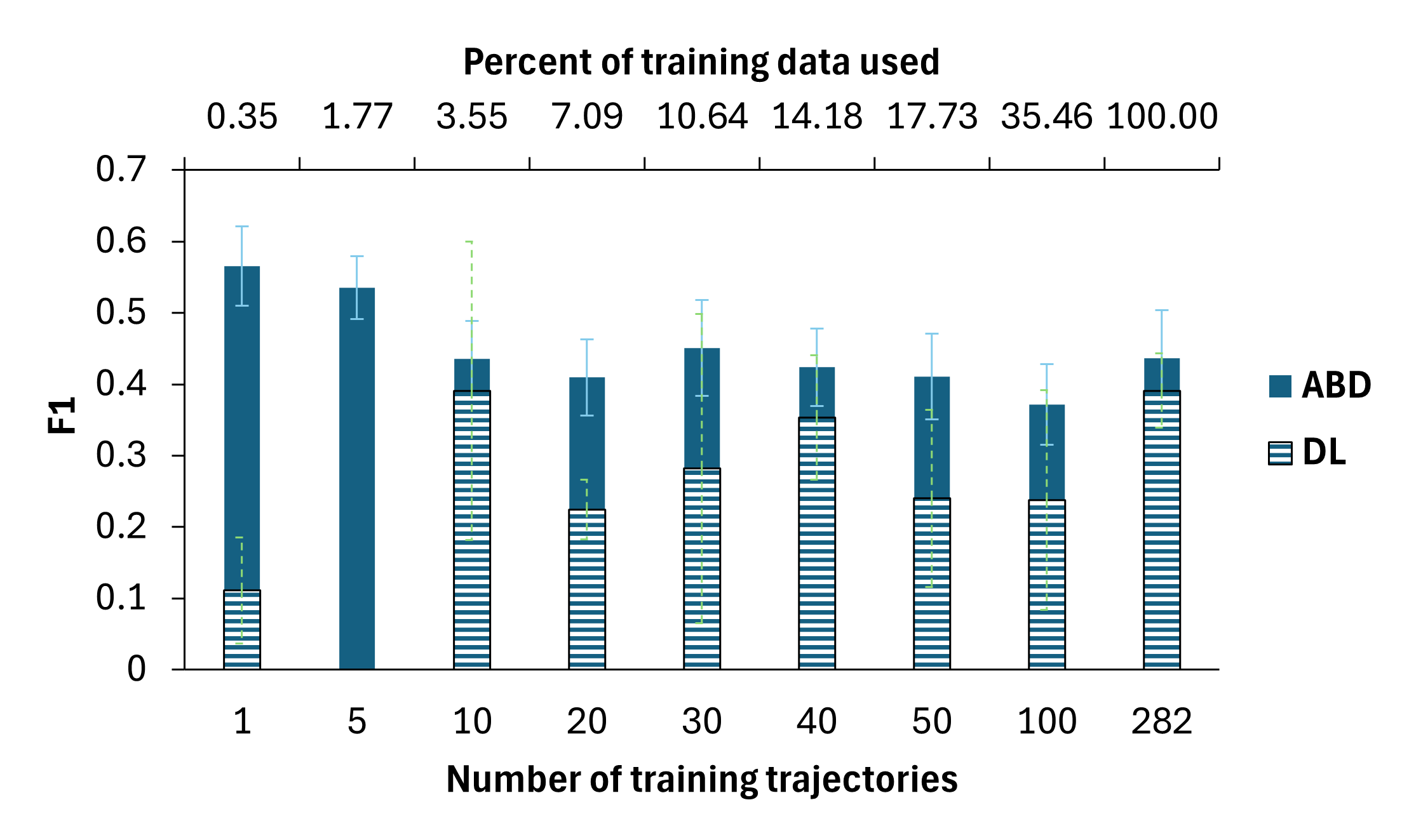}
  
    \caption{ Evaluation of ABD and DL with limited training data. DL achieved a zero F1 with five training trajectories.}
    \label{fig:dleff}
    \vspace{-10pt}
\end{figure}

\noindent\textbf{Data Efficiency.} The ABD model also works well with limited training trajectories as seen in Figure~\ref{fig:eff} while DL-based methods are more data-driven as seen in Figure~\ref{fig:dleff}. Note that for ABD, the use of a single training trajectory versus all of the historical data gave the same precision of $0.62$ and a $0.13$ difference in F1. On the other hand, as expected, DL has a huge variation with increasing training trajectories by boosting its performance by $254\%$ as seen in Figure~\ref{fig:dleff}. This gives scope for the application of our model with expensive or limited available data. 

\begin{figure}[]
  \centering
  \includegraphics[scale=.42]{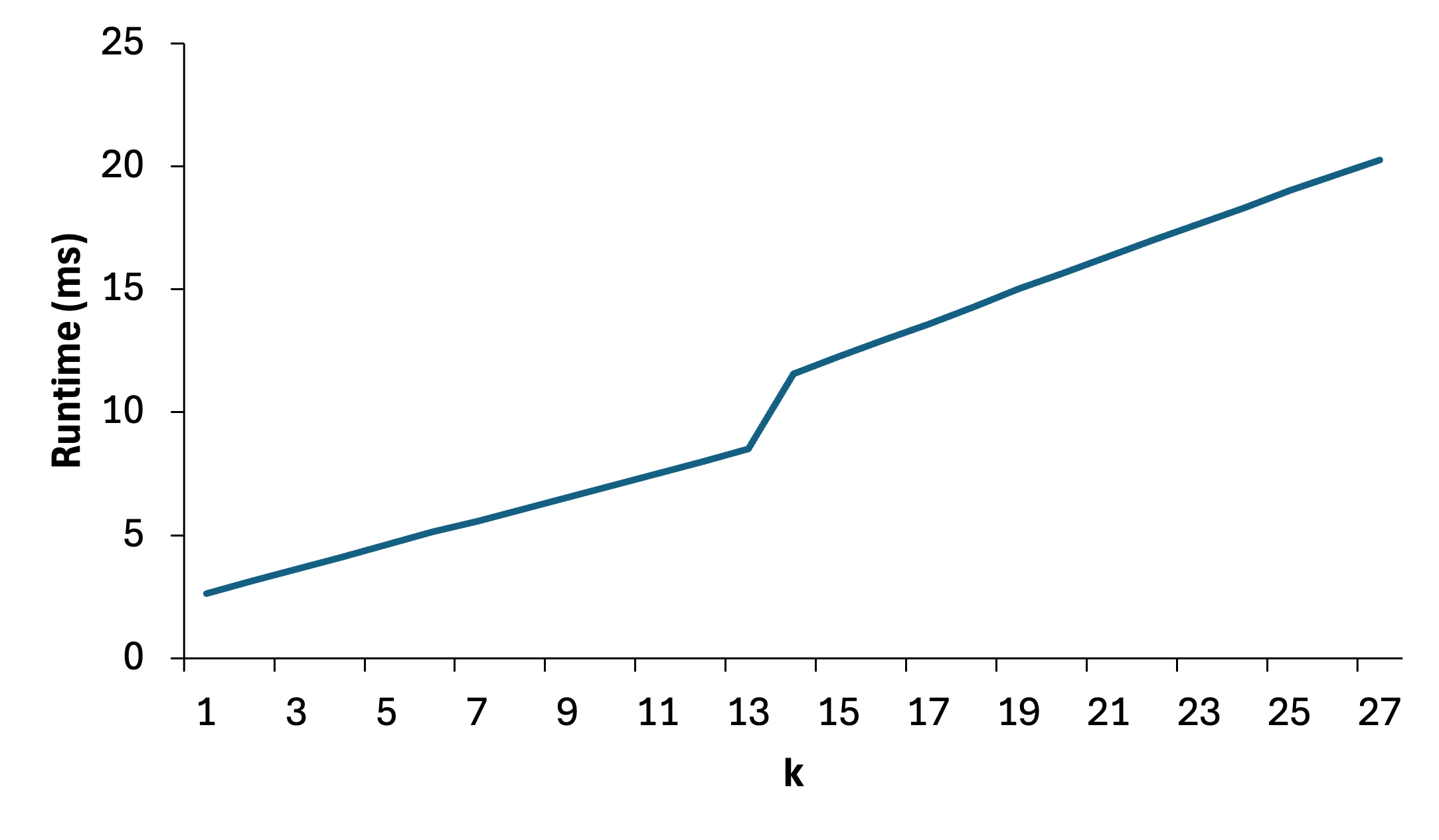}
  \vspace{-3pt}
    \caption{ Evaluation of runtime in terms of milliseconds of ABD as a function of $k$.}
    \label{fig:runtime}
    \vspace{-8pt}
\end{figure}

\noindent\textbf{Runtime.} In Figure~\ref{fig:runtime} we examine the runtime of ABD as a function of the number of regions $k$.  As expected, the runtime increases linearly with $k$ as this simply involves additional deductive steps.  Further, we note that the deduction itself is efficient (linear in the size of nodes) as previously reported~\cite{aditya2023pyreason,ssTAI22}.

\noindent\textbf{Explainability.} All regions are symbolic in nature, every inference can be backtracked to the sequence of historically learned rules, in addition to its confidence as seen in Figure~\ref{fig:xai}. This gives scope for domain experts in analyzing false predictions, assess vessel behavior, and even incorporating domain knowledge into the rules.

\section{Deployment}
\label{sec:deploy}
\begin{figure}[t]
  \centering
  \includegraphics[scale=.38]{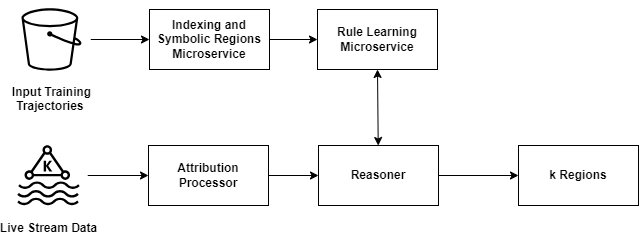}
  \caption{Deployment of abduction model in an online learning setting with kafka.}
        \label{fig:depl}
        \vspace{-8pt}
\end{figure}

We designed a prototype system based on the abduction model with a live feed of trajectories, where it continuously updates its logic program as it generates regions in an online learning setting. This architecture is depicted in Figure~\ref{fig:depl}.  We use a microservices-based architecture for near real-time detection of maritime dark vessels that receives input training data delivered by data providers to an Amazon S3 bucket. The arrival of new data triggers a batch process that performs data indexing and generates symbolic regions. This processed data is then fed into a rule-learning microservice, which is subsequently transformed into a logic program by learning rules that are staged into the S3 bucket.
In the production environment, live data is streamed via a Kafka feed. We use Apache Kafka to consume the AIS data stream in near real-time as a streaming architecture. An attribution processor subscribes to this feed and enriches the incoming data by tagging it with the necessary regions and indexing metadata. The enriched data is then integrated into the logic program (which includes both updated rules and TAFs), before being fed into the reasoner ($\Gamma^*$), which infers $k$ regions at a given time horizon. We then use Quantum Geographic Information System software to visualize the regions in the AOI for an end-user.

\section{Conclusion}
We identify the locations of dark maritime vessels using a combination of abductive inference and rule learning and provides explainable long-time horizon prediction - an area where machine learning approaches fail.  These aspects were validated by our experimental results and we provide our deployment architecture with a live feed of data.
This work can be extended by leveraging environmental knowledge in the logic program, which has a significant role in the maritime domain where we look to utilize techniques from neurosymbolic AI~\cite{shakarian2023neuro} that will enable the use of larger scale models for enhanced near-term precision while retaining the long-term reasoning ability of the abduction methods introduced in this paper. One direction for future work is to examine the case where an adversary is taking action to reduce it's ability of detection by algorithms such as those presented in this paper. For example, extending our work in the paradigm of adversarial geospatial abduction~\cite{DBLP:journals/tist/ShakarianDS12} can take us in this direction.

%





\begin{acks}
This work was funded by ONR grant N00014-23-1-2580 and used maritime data provided by Spire.  There are pending patent(s) on ideas presented in this paper including U.S. provisional 63/745,753. We thank Priyank Vyas for helping us in engineering related tasks for the experiments.
\end{acks}




\bibliographystyle{AAMAS-2025-template/ACM-Reference-Format} 
\bibliography{main.bib}

\appendix

\section{Appendix}{\bf Technical Preliminaries}

\noindent\textbf{Logical Language and Syntax.} The language is defined with a set of constants that is partitioned into multiple domains ($\domainSet_i \subset \constantSet$),
examples of such subsets are: $l_1, l_2,.. \in \domainSet_{loc}$ - a set of all potential locations of the vessel, $\regionconst_{l1,l2},.. \in \domainSet_{\regionconst}$ - a set of all regions, and $agt,.. \in \domainSet_{agt}$ - a set of agents, as specified in Section~\ref{subsec:prelims}.
As usual in first-order logic, we define a corresponding set of variables ($\variableSet$), and a set of predicate symbols ($\predicateSet$) with an 
arity (notated by $arity(p) \in \mathbb{W} $, where $p \in  \predicateSet$ and $\mathbb{W}$ is a set of all whole numbers).
Terms (constants or variables) and predicates form atoms. Atoms and their negation are called literals. Literals are called ground literals if and only if there is no occurrence of any variables (they are replaced by elements from $\constantSet$). Consider $\groundLiteralSet$ to be the set of all ground literals. A formula can be defined recursively- any literal is a formula and any combination of two formulae with logical connectives (conjunction $\wedge$, disjunction $\vee$ and negation $\neg$) is a formula.

\noindent\textbf{Annotated Language.} In addition to the first-order logic syntax and semantics, we allow for literals to be annotated with elements (intervals in $[0,1]$) of a lower semi-lattice structure  $\semiLattice$ (not necessarily complete) with ordering $\sqsubseteq$ \cite{ks92,ssTAI22}. 
We assume the existence of a set $\avar$ of variable symbols ranging over $\semiLattice$, and $h$ is an $n$-ary function symbol over $\semiLattice$ and $\mu_1,\ldots,\mu_n$ are annotations, then annotations can be defined~\cite{ssTAI22, ks92} as any member of $\semiLattice \cup \avar$ or $h(\mu_1,\ldots,\mu_n)$.
\begin{figure*}[!t]
    \begin{center}
        \includegraphics[width=.8\linewidth]{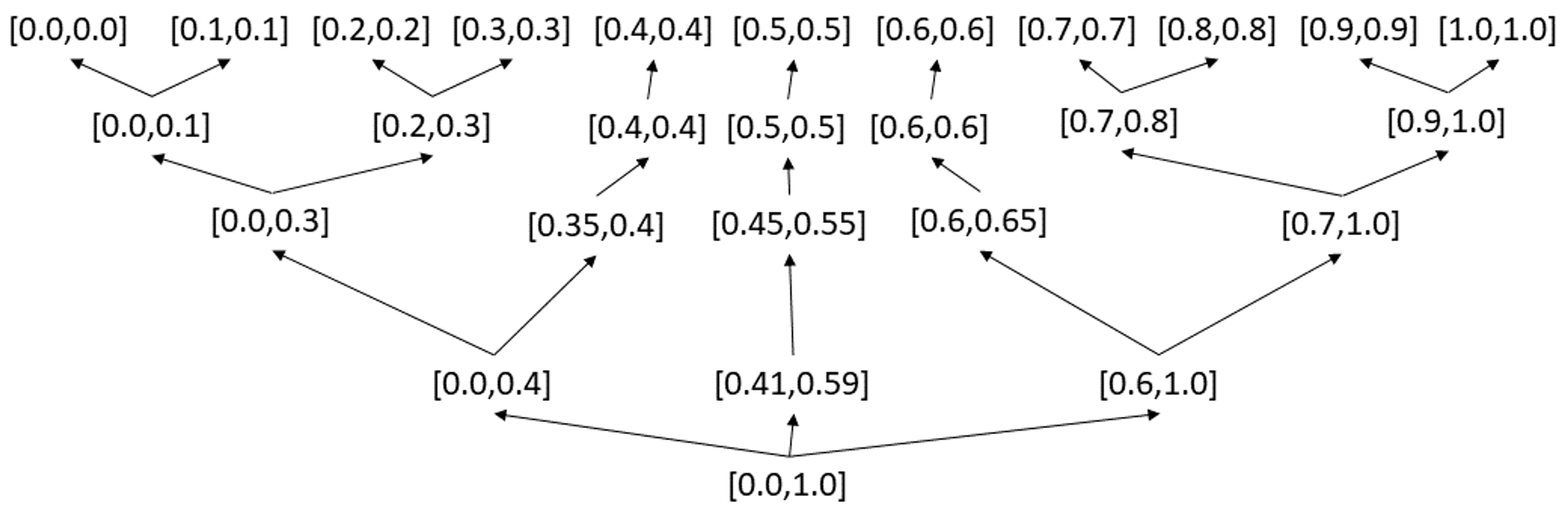}
    \end{center}
    \caption{\label{fig:lowerLattice}Example of a lower semi-lattice structure where the elements are intervals in $[0,1]$.}
\end{figure*}

 $\semiLattice$ has a single bottom element ($\bot$) and a set of top elements $\top_{0} = [0,0]$, . . . $\top_{i}$ . . . $\top_{max} = [1,1]$. The notation height ($\top$) is the maximum number of elements in the lattice in a path between $\bot$ and a top element. The annotations are represented with $[\ell,u]$ which is simply a subset of the unit interval $[0,1]$ - which generalizes both fuzzy and classical logic. 
As seen in the Figure~\ref{fig:lowerLattice}, we set the bottom element to be $\bot = [0,1]$ and define a set of top elements $\{[x,x] \; | \; [x,x]\subseteq [0,1]\}$.  Note that the notation ``$\sqsubseteq$'' in a semi-lattice of bounds,  has the following intuition: $[0,1] \sqsubseteq [1,1]$ in this case.
We write an annotated literal $a:[\ell,u]$ to mean that the literal $a$ has truth value associated with interval $[\ell,u]$. We refer the reader to \cite{ks92,ssTAI22} for lattice-theory justification of this approach and how it generalizes other logical paradigms. 
We also note that we have learned our logic programs in a way to treat these bounds as confidence (see Section~\ref{sec:rule_learning} in the main paper).

\noindent\textbf{Extensions to Temporal Syntax.}  We follow the extension of temporal syntax and semantics~\cite{aditya2023pyreason,bavikadi2024geospatial,shakAamas13} to form temporally annotated facts (TAFs) and annotated formulae. For an annotated literal $f$ that is true at time $t$, $f_t$ is a TAF. Annotated formulae are constructs formed with operators like $\after(f,f')$ as well as annotated literals. For annotated literals $f,f'$, $\after(f,f')$ is interpreted as $f$ occurs after $f'$. Note that TAFs are considered separately from these formulas when we describe the semantics.

\noindent\textbf{Semantics and Logic Program.} As per previous work on temporal annotated logic~\cite{shakAamas13,aditya2023pyreason,bavikadi2024geospatial}, given a set of timepoints $T$, a set of all (ground) literals $\groundLiteralSet$, an interpretation $\interpretation$ is any mapping $\groundLiteralSet \times T \to \semiLattice$. 
The set~$\interpretationSet$ of all interpretations can be partially ordered via the ordering: $\interpretation_1\preceq \interpretation_2$ if and only if for all ground atoms $g \in \groundLiteralSet$ and time $t \in T$, $\interpretation_1(g, t)\sqsubseteq \interpretation_2(g, t)$. $\interpretationSet$ forms a complete lattice under the $\preceq$ ordering. We define a satisfaction relationship ``$\models$'' and rules for temporally annotated extensions as defined in~\cite{shakAamas13, aditya2023pyreason}.  
We read $ \interpretation \satisfactionAtTime{t} \groundedLiteral:\mu$ as an interpretation $\interpretation$ \emph{satisfies} an annotated ground literal $\groundedLiteral:\mu$ where $\groundedLiteral \in \groundLiteralSet$ at time $t$, iff $\mu \sqsubseteq \interpretation( \groundedLiteral, t)$.
Given TAF $f_t$ and interpretation $I$, the interpretation $I$ satisfies the TAF $f_t$, iff $I \models_t f$. 
If $\after( p_1,p_2): \mu$ is an annotated formulae for unary predicate symbols $p_1,p_2 \in \predicateSet$, then there exists $r_1,r_2 \in \domSet_{r}$ and $agt \in \domainSet_{agt}$ such that $I \models_{t} \atPred(agt,r_1): \mu$, $I \models_{t'} \atPred(agt,r_2): \mu$ (where $t' < t$),  $I \models_{t} p_1(r_1):\mu$, and $I \models_{t'} p_2(r_2):\mu$.

A program $\Pi$ is a set of rules, where each has an annotated atom in the head and a conjunction of annotated formulae in the body. Sample rules in $\Pi$ are included in Table~\ref{tab:examplerlsRules}. A rule with no body is called a fact. A rule with no variables is called a ground rule. If an interpretation $\interpretation$ satisfies every annotated formula in the body of $r \in \Pi$, then the annotated atom in the head of $r$ must be satisfied by $\interpretation$. An interpretation $\interpretation$ is said to satisfy $r$, if and only if it satisfies all ground instances of $r$. An interpretation $\interpretation$ is said to satisfy $\Pi$, if and only if $\interpretation$ satisfies every rule and TAF in $\Pi$.  The minimal model is an interpretation that can be thought of everything that can be concluded from deductive inference and commonly used for entailment queries in annotated logic~\cite{ks92,shakAamas13,ssTAI22,aditya2023pyreason}.  This is often computed using a fixpoint operator as done in the aforementioned work - and refer the reader to the well-established work on that topic for details.  In this work, we slightly abuse the notation of \cite{ks92} and use $\Gamma^*(\Pi)$ to denote the minimal model of $\Pi$. Here, consider $\Gamma$ to be a function that returns a model of a given $\Pi$ and $\Gamma*$ to be a convergence of $\Gamma$ over multiple applications, that returns the minimal model.

\noindent\textbf{Abductive Inference Framework}
A specific logic-based abduction can be formally described as: \\
Consider a logical theory $\program$ which encodes a particular domain, a set $O$ for observations described by a set of atomic formulas, and a
set $H$ of (usually atomic) formulas containing possible individual hypotheses,
find an explanation (or solution) for $O$, that is, a suitable set $E  \subset  H$ such that
$\program \cup E$ is consistent and  $\program \cup E \models O$. 

For our use-case, for a given an agent $agt$, initial conditions $\Pi_{init}$, behavioral rules $\Pi_{behav}$, and ground-truth trajectory $\tau_{agt}$, \\ $\langle agt, \Pi_{init}, \Pi_{behav}, \tau_{agt} \rangle$ is the abduction problem. The goal is to find an explanation $\Pi_{pred}$ for $\langle agt, \Pi_{init}, \Pi_{behav}, \tau_{agt} \rangle$.

Here, we consider $\Pi_{pred}$ to be $E$, $\Pi_{behav} \cup \Pi_{init}$ to be $\program$, and  $\tau_{agt}$ to be $O$. As seen in Section~\ref{sec:abduc}, $\program \cup E$ is consistent and  $\program \cup E \models O$. The intuition behind the latter is from the transitive property of entailment [For any formula or a set of formulae, say $A, B, C$, if $ A \models B$, $ B \models C$, then $ A \models C$.], as $\Pi_{behav} \cup \Pi_{init} \models \Pi_{pred}$ and $\Pi_{pred} \models \tau_{agt}$.

\section{Appendix}{\bf Additional Experiments with external ground truth data}
\label{app:dgv}

While we are not aware of any academic work that uses ground truth for actual dark vessels – the setup we proposed is adapted our use-case from the prior work in the area~\cite{dlbaseline,dstnet,txlstm, tf, tf2, tp}, where they mask the historical trajectory to extract input and ground truth.  This limitation is likely due to the requirement for non-AIS data to conform with the ship location. That said, our industry partner has provided us with several exemplars where they had confirmation based on an external source where they have verified AIS tampering.  They have provided us four such trajectories.  We did an experiment with this data and following the table~\ref{app:tab:dgv} of the area needed to cover the tagged dark activity along with fraction of area of top-k regions needed to cover the doppler tagged dark activity out of the search space for four test cases.  Here the model was trained in the normal way (with standard trajectories). 

\begin{table}[h!]

\begin{tabular}{|l|l|}

\hline
\textbf{Test Sample} & \textbf{Area in $Km^2$ (Percent of possible area)} \\ \hline
1                    & 30.11 (11)                                                        \\ \hline
2                    & 28.70 (9)                                                         \\ \hline
3                    & 12.64 (4)                                                         \\ \hline
4                    & 54.41 (17)                                                        \\ \hline
\end{tabular}
\caption{Coverage of tagged dark activity: Proportion of area required to encompass Doppler-tagged dark activity and fraction of area from top-k regions within the search space across four test cases.}
\label{app:tab:dgv}
\end{table}


\end{document}